\documentclass[conference]{IEEEtran}
\IEEEoverridecommandlockouts
\usepackage{cite}
\usepackage{amsmath,amsfonts}
\usepackage{amssymb}
\usepackage{physics}
\usepackage{algorithm}
\usepackage{algpseudocode}
\usepackage{multirow}
\usepackage{graphicx}
\usepackage{textcomp}
\usepackage{xcolor}
\usepackage{float}
\renewcommand{\baselinestretch}{.99}
\algnewcommand\algorithmicforeach{\textbf{For Each:}}
\algnewcommand\ForEach{\item[ \algorithmicforeach]}
\def\BibTeX{{\rm B\kern-.05em{\sc i\kern-.025em b}\kern-.08em
    T\kern-.1667em\lower.7ex\hbox{E}\kern-.125emX}}
    
\begin{document}
\title{SafeNav: Safe Path Navigation using Landmark Based Localization in a GPS-denied Environment
}
% \author{
%     \IEEEauthorblockN{Ganesh Sapkota,
%                       Sanjay Madria
%                      }
%     \IEEEauthorblockA{Department of Computer Science,
% Missouri University of Science and Technology, Rolla, MO, USA\\
% Email: \{gsapkota, madrias\}@mst.edu}
% }

\author{\IEEEauthorblockN{Ganesh Sapkota}
\IEEEauthorblockA{\textit{Department of Computer Science} \\
\textit{Missouri University of Science and Technology, USA}\\
gs37r@mst.edu}
\and
\IEEEauthorblockN{Sanjay Madria}
\IEEEauthorblockA{\textit{Department of Computer Science} \\
\textit{Missouri University of Science and Technology, USA}\\
madrias@mst.edu}
}

% \author{
% % \IEEEauthorblockN{XXX}
% % \IEEEauthorblockA{\textit{Department of Computer Science} \\
% % \textit{XXX University}\\
% % XXX Location \\
% % XX email}
% \and
% % \IEEEauthorblockN{YYY}
% % \IEEEauthorblockA{\textit{Department of Computer Science} \\
% % \textit{XXX University}\\
% % XX location\\
% % XX email}
% }
\maketitle
%------------------Each section are in sections folder----------------------
% \input{sections/1_abstract}    
% \input{sections/2_introduction}
% \input{sections/3_related_works}
% \input{sections/4_system_model}
% \input{sections/5_experiments_results}
% \input{sections/6_conclusion_futureworks}
% \input{sections/7_references}
%%%%%%%%%%%%%%Abstract%%%%%%%%%%%%%%%%%%%%%%%%%%%%%%%%%%%%%%%%%%%%%%%%%%%%%%%%%%

\begin{abstract}
In a battlefield, %the reliance on Global Positioning System (GPS) for navigation can be a critical vulnerability. 
adversaries often disrupt or deceive GPS signals, necessitating alternative methods for the localization and safe navigation of troops. Vision-based localization approaches like Simultaneous Localization and Mapping (SLAM) and Visual Odometry (VO) use sensor fusion and are computationally intensive. Range-free localization methods such as DV-HOP suffer from accuracy and stability in dynamic and sparse network topology. 
Using a landmark-based localization (LanBLoc) and battlefield-specific motion model (BMM) aided with an Extended Kalman Filter (EKF), this paper first proposes a navigation approach called LanBLoc-BMM. The efficiency of the proposed approach was compared with three other state-of-the-art (SOTA) visual localization algorithms integrated with BMM and Bayesian filters EKF \& PF for navigation on both synthetic and real trajectory datasets. LanBLoc-BMM (with EKF) shows the best accuracy in terms of Average Displacement Error (ADE) and Final Displacement error (FDE) of 0.2393 and 0.3095, respectively, AWRS of 0.1528 and runtime of 14.94ms on real-dataset, among all SOTA algorithms. %Using the centroid method, an average trajectory length error of  5.46\%, ADE/FDE of 1.11/0.99 m, and AWRS of 1.01 were observed when EKF is integrated which is 16.12\%  of improvement in trajectory length error 1.67 times better than the ADE, 3.30 times better than the FDE of the Convex-Hull based approach.
Then, two different safe path navigation algorithms called SafeNav-CHull and SafeNav-Centroid are proposed by integrating LanBLoc-BMM( EKF). The SafeNav-Centroid relies on the centroid of the safe path segments, while SafeNav-CHull uses the convex hull of the path segments for movement decisions. Both algorithms integrate a novel Risk-Aware RRT*(RAw-RRT*) algorithm for obstacle avoidance and risk exposure minimization. A new evaluation metric called Average Weighted Risk Score (AWRS) is introduced to evaluate the risk exposure of the trajectory.
We presented a simulated battlefield scenario for safe path navigation and evaluated the estimated trajectories of proposed safe path navigation algorithms regarding displacement errors, risk exposure, computation time, and trajectory length.
The SafeNav-Centroid shows efficient performances in terms of ADE, FDE, AWRS, and Average percent error, while SafeNav-CHull presents the best computation time performance.
% The results demonstrate that our approach not only ensures the safety of the mobile units by keeping them always within the safe path but also enhances operational effectiveness by adapting to the evolving threat landscape. 
\end{abstract}

%%%%%%%%%%%%%%%%%%%%%%%%%%%%%%%%%%%%%%%%%%%%%%%%%%%%%%%%%%%%%%%%%%%%%%%%%%%%
%%%%%%%%%%%%%%%%%%%%%%%%%%%%%Introduction%%%%%%%%%%%%%%%%%%%%%%%%%%%%%%%%%%%
\section{Introduction}
% \red{Red lines indicate Comments} and \blue{BLue lines indicate Updated contents}\\
In military operations, %the secure navigation of mobile troops on a battlefield is a critical and complex task. 
safe navigation % on a battlefield is a multifaceted concept integrating advanced technological solutions, strategic resource management, and comprehensive security measures to address various operational challenges. It 
demands a precise and reliable positioning system\cite{positioning_1}, path planning, and maneuver decision support systems under adversarial conditions. Thus, positioning algorithms are central to safe navigation, especially on a battlefield, where precise location, timing, and coordination are critical for moving troops and ground vehicles. %These algorithms determine the position of moving troops, ground vehicles, or any asset, often in challenging and dynamic environments. 
In scenarios where adversarial threats like GPS jamming, terrain obstacles, and hazards are prevalent, robust navigation systems play a crucial role in mission planning and maintaining operational effectiveness. Moreover, as autonomous systems like unmanned ground vehicles (UGVs) and unmanned aerial vehicles (UAVs) are becoming integral to modern warfare, the demand for accurate and resilient navigation capabilities continues to grow. These systems ensure the safety, timing, and coordination for successful battlefield maneuvers and mission execution.

{\textbf{Challenges: }}GPS provides reliable location data however, its dependency on satellite signals makes it vulnerable to jamming and spoofing in combat scenarios\cite{gps_jamming}. Several methodologies have been developed to address localization and navigation challenges for GPS-Denied environments, including, Simultaneous Localization and Mapping (SLAM) \cite{ref_slam}, Visual Odometry (VO) \cite{vo_review}, and other non-GPS-based techniques like DV-Hop Localization \cite{range_free_localization}. SLAM technique, where GPS is either unreliable or unavailable, involves the creation of a map of an unknown environment while simultaneously keeping track of the individual's or vehicle's location within that environment. VO estimates the position and orientation of a moving object by tracking and analyzing the movement of visual features (such as edges or corners) between consecutive frames of video captured by onboard cameras. While SLAM and VO are powerful techniques for pose estimation and navigation, especially in GPS-denied environments, they come with certain limitations, particularly in safe battlefield navigation. These algorithms require significant power to process high-resolution data or complex 3D environments, which is computationally intensive %. power\cite{slam_survey}, 
and might be a constraint in smaller, battery-powered, or rapid-response systems. Also, they require multiple sensor fusion techniques\cite{sensor_fusion}, which makes it not feasible in resource-constrained settings like battlefields where soldiers use small and battery-powered mobile devices.

To overcome the challenges and limitations given above, \cite{lanbloc_wowmom} introduced LanBLoc, a localization framework designed for non-GPS battlefield environments. It utilizes stereo vision and landmark recognition, relying solely on passive camera sensors and geographical landmarks as reference points for localization. However, their approach does not tackle navigation and motion planning in non-GPS environments using landmark-based localization, which is the main focus of our work here.

{\textbf{Our Contributions: }} 
\begin{figure}[!ht]
\centering
\includegraphics[width=0.45\textwidth]{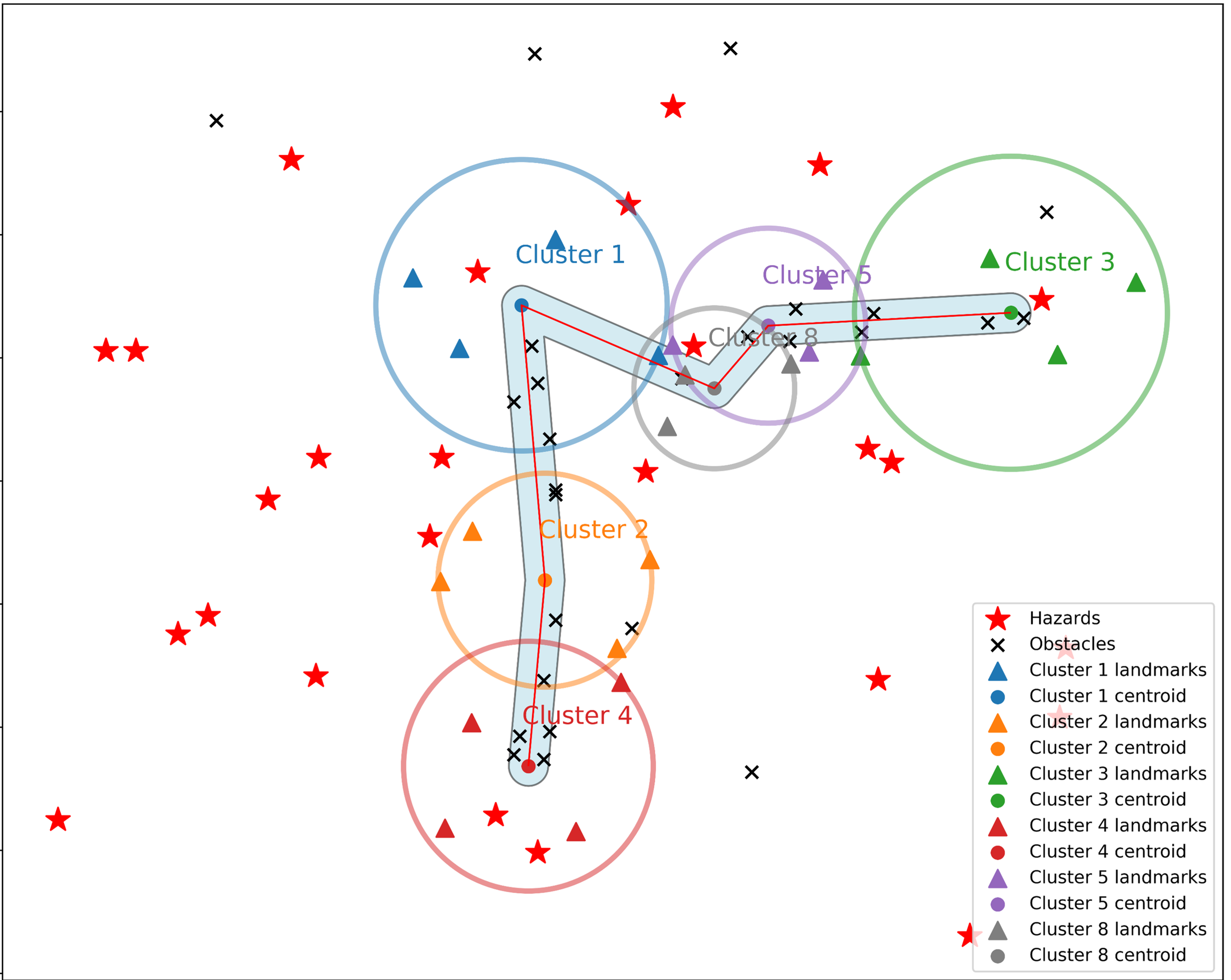}
\caption{A battlefield environment cluttered with landmark clusters, obstacles, and hazard components. The safe path buffer starting from centroid cluster 4 and ending at the centroid of cluster 3 indicates a risk-free zone, guides the moving entity safely in avoiding obstacles, and minimizes risk exposure.}
\label{battlefield_map}
\end{figure}
This research primarily focuses on developing solutions to efficiently guide moving military troops, ground vehicles, or reconnaissance units (considered as Moving entities (ME)) along a \textit{safe path} in a non-GPS environment using LanBLoc as a location measurement model. The \textit{safe path} refers to a navigational boundary in the battlefield that minimizes the risk exposure due to hazardous components, ensuring the safety, efficiency, and successful navigation of a  ME from source to the destination, as illustrated in Fig.\ref{battlefield_map}. 

To achieve our objective, %we first performed extensive research to evaluate the efficiency of the LanBLoc framework for ground navigation applications, considering a ground scenario with real-world landmark data.
we first define motion planning of moving objects for ground movement and then adapt it for battlefield scenarios called \textbf{Battlefield Motion Model (BMM)} that captures the essential non-linear dynamics of battlefield terrain, such as obstacles and hazards.
Then, we propose a navigation approach \textbf{LanBLoc-BMM}  using the LanBLoc as a location measurement model and BMM for motion planning, integrating Extended Kalman Filter (EKF) \cite{ekf_2} and Particle Filter (PF) \cite{pf2} separately to handle unmodeled influences of battlefield terrain.
The performance of our proposed LanBLoc-BMM approach is evaluated and compared with state-of-the-art (SOTA) visual localization methods from literature such as DeepLoc-GPS, DeepLoc-EKF-GPS \cite{deepLocalization}, Multiple Sensor Fusion %(IMU/OD/VI/MA) 
\cite{sensor_fusion}, using synthetic and real trajectory datasets. Experimental results showed that the LanBLoc-BMM approach, when integrated with EKF, produces the best navigation performance for both synthetic and real trajectory datasets, demonstrating the effective application of the LanBLoc framework as a location measurement model in battlefield navigation applications.
% To achieve our primary goal, we developed an efficient navigation approach combining BMM \& LanBLoc and integrating EKF \cite{ekf_2} to handle unmodeled influences of battlefield terrain.
Thus, our LanBLoc-BMM(EKF) based navigation approach helps to navigate the moving entity in a GPS-denied battlefield environment from the source to the target destination, as shown in Fig \ref{battlefield_map}. The proposed navigation approach assumes that the battlefield environment offers sufficient landmarks within the detectable range of moving objects to facilitate trilateration, like in \cite{lanbloc_wowmom}. In the figure, circular shapes indicate the landmark clusters, and the bounded region(strip) starting from the centroid of cluster 4 and ending at cluster 3 indicates the safe path. 

Additionally, to ensure the safety of the moving entity, we then proposed two separate algorithms called \textbf{SafeNav-CHull} and \textbf{SafeNav-Centroid} that ensure safety checks of the next possible state and controlled maneuvering within the safe path boundary.
%using convex hull (CHull) \cite{convex_hull_app1} and centroid properties of path segments. 
Each algorithm is then integrated with a LanBLoc-BMM(EKF) for safe and efficient navigation along the given safe path. To handle the obstacles within the safe path, we also proposed a new variant of the Rapidly Exploring Random Tree-Star (RRT*) algorithm called \textbf{Risk-Aware RRT* (RAw-RRT*)} by defining a novel risk-aware cost function.
The SafeNav-Centroid computes the centroids of the given safe path segments for controlled maneuvering of the units, ensuring safety and the shortest trajectory. On the other hand, the SafeNav-CHull applies a point-in-polygon test \cite{even_odd_algo} on the convex hull \cite{convex_hull_app2} of safe path segments to ensure the safety and control input calculation. RAw-RRT* ensures an obstacle-free path with the shortest length and minimal risk of hazards. We evaluated the performance of the proposed algorithms using standard metrics such as Average Displacement Errors(ADE), Final Displacement Error, and Percent Error. Additionally, to evaluate the safety, we defined a metric called \textbf{Average weighted Risk Score (AWRS)}, which measures the average risk exposure of trajectory generated by navigation algorithms by taking account of the deviation of trajectory points from the hazard boundary.
The performance contributions of our research work are:

% \begin{enumerate}
\textbf{1)} We evaluate and compare the navigation efficiency of the LanBLoc-BMM algorithm with three other SOTA localization algorithms integrated with EKF and PF for navigation on both synthetic and real trajectory datasets. We observe the best accuracy \(ADE / FDE\) of 0.4972 / 0.4675, minimal AWRS of 0.3492, and minimal runtime of 3.98ms on synthetic datasets and achieved (\(ADE / FDE\) of 0.2393 / 0.3095), minimal AWRS of 0.1528   and the least runtime of 14.94 ms for LanBLoc-BMM (EKF) on a real dataset. 

\textbf{2)} We compare the performance of proposed SafeNav-CHull and SafeNav-Centroid algorithms, both integrated with RAw-RRT*.
    %an obstacle avoidance algorithm called RAw-RRT*that guarantees the safest, shortest, and obstacle-free route along a predetermined safe while navigating towards the destination. 
We obtain an average percent error in trajectory estimation of 5.46\% for the SafeNav-Centroid algorithm which is a 16.12\% improvement over the SafeNav-CHull approach. We obtain an average ADE/FDE of 1.11/0.99m, for the SafeNav-Centroid algorithm, which is 1.67 times and 3.30 times better, respectively, than the ADE and FDE of the SafeNav-CHull algorithm. SafeNav-Centroid also presents lower AWRS compared to SafeNav-Chull; however, SafeNav-Chull demonstrates higher efficiency in terms of computational time.
Overall, SaveNav-Centroid generates the shortest path with less exposure to hazard, while SafeNav-Chull provides comparable results with higher computational efficiency.
% \end{enumerate}
% This paper is organized into five Sections. Section I introduces the research problem, and motivation then highlights our contributions to solving the problem. Section II discusses Related Works. Section III describes our approach and provides an overview of the system in detail. Section IV discusses the experimental setup, dataset, evaluation metrics, performance results, and comparison with existing works. Section V concludes the paper with the application and future direction of research.
%%%%%%%%%%%%%%%%%%%%%%%%%%%%%%%%%%%%%%%%%%%%%%%%%%%%%%%%%%%%%%%%%%%%%%%%%%%%
%%%%%%%%%%%%%%%%%%%%%%%%%%%%%%%%Related Works%%%%%%%%%%%%%%%%%%%%%%%%%%%%%%%
\section{Related Works}
%%%%%%%%%%%%%%%%%%%%%%%%%%%%%%%%%%%%%%%%%%%%%%%%%%%%%%%%%%%%%%%%%%%%%%%%%%%%%%%%%%%%%%%%%
The recent advancements in GPS-free positioning, including SLAM and VO, have significantly improved the navigation and mapping capabilities of autonomous systems and unmanned ground vehicles. %, particularly in challenging GPS-denied environments.
%%%%%%%%%%%%%%%%%%%%%%%Vision-based methods%%%%%%%%%%%%%%%%%%%%%%
Vision-based methods\cite{ref_slam} %, for robot localization, have achieved much attention in recent years due to their ability to
provide precise and reliable localization results in GPS-denied environments. They utilize image and video data to determine the robot's location within its surroundings. SLAM technique ~\cite{ref_slam} estimates the position and orientation $(pose)$ of a robot within an environment while simultaneously creating its map. However, this technique requires a fusion of multiple cameras, lidar, or radar sensors to generate a map of its surroundings. VO methods \cite{deepLocalization} also estimate the robot's motion by tracking the changes in images or video frames captured by the robot. 
%Structure-from-Motion algorithm (SfM)\cite{ref_sfm} reconstructs the 3D structure of the scene from multiple images, which is useful for outdoor environments with well-defined features.
%%%%%%%%%%Deep Learning-based Localization%%%%%%%%%%%%%%%%%%%%%%%%%%
Deep Learning-based Localization\cite{ref_pose_estimate_yolo}, particularly convolutional neural networks (CNNs), have demonstrated impressive performance in vision-based localization %. CNNs can extract high-level features from images and videos, enabling robust localization.
in various environments.
%%%%%%%%%%%%%%%%%%DV-Hop Methods%%%%%%%%%%%%%%%%%%%%%%%%%%%%%%%%%%%
DV-hop-based range-free localization algorithms commonly used in wireless sensor networks $(WSNs)$\cite{dv_hop_Cyclotomic} have received enormous attention due to their low-cost hardware requirements for nodes and simplicity in implementation. The algorithm estimates the location of unknown nodes based on their hop count to known anchor nodes and the known location of the anchor nodes. However, these methods only perform well in static, isotropic, and uniformly distributed network topology. They are sensitive to node mobility and cannot adapt to dynamic network conditions.
%%%%%%%%%%%%%%%%%%%%%%%Baysian Filters%%%%%%%%%%%%%%%%%%%%%%%%%%%%%%%
%For linear systems with Gaussian noise, the 
% Kalman Filter(KF) provides a framework for predicting future states of moving objects and estimating the current state. It is computationally efficient for real-time applications. However, KF is only optimal for linear systems whose performance degrades with non-linearity. 
%The non-linear systems are better handled by other Bayesian filters such as Extended Kalman filter $(EKF)$ \cite{ekf_2} and Particle filter$(PF)$. 
% The Kalman Filter (KF) efficiently predicts future states and estimates current states in real time but is optimal only for linear systems, with performance degrading in non-linear scenarios.
EKF \cite{ekf_2} extends the Kalman filter to handle nonlinear systems by linearizing them at each time step by calculating Jacobians of estimating functions. %It is more popular in many applications, especially in robotics and aerospace, for tasks like localization and tracking\cite{ekf_2}. 
PFs though suitable for non-Gaussian and highly nonlinear systems\cite{pf2} can be computationally expensive if the right number of particles is not selected. So, careful tuning of the right number of particles and the resampling strategy is required, which can eventually affect the filter's performance. 
%UKF also handles linearization by using a deterministic sampling approach, providing a more accurate approximation for nonlinear systems. UKF can handle arbitrary noise distributions more effectively than EKF. However, UKF is more computationally intensive than EKF due to the sigma point calculation and propagation. Selecting appropriate parameters for the sigma points can be challenging and may require empirical tuning.
%Among the other filters, the Extended Kalman Filter $(EKF)$ is particularly feasible and efficient for use in battlefield scenarios due to its ability to handle non-linear systems, which are commonly encountered in military applications. 
The $EKF$ is effective for battlefield scenarios due to its balance of accuracy, computational efficiency, robustness, and ability to handle non-linear systems, making it widely used in military applications.

 \section{Safe Path Navigation Framework}
%%%%%%%%%%%%%%%%%%%%%%%%%%%%%%%%%%%%%%%%%%%%%%%%%%%%%%%%%%%%%%%%%%%%%%%%%%%%%%%%%%%%%%%%%
% \subsection {System Overview}
Our system is designed to guide a mobile troop through an obstacle and hazard-free trajectory starting from its current position to the destination. The moving entity (node) is required to navigate within the safe path to ensure the safety, survivability, and success of the mission.

\begin{figure}[!ht]
\centering
\includegraphics[width=0.46\textwidth]{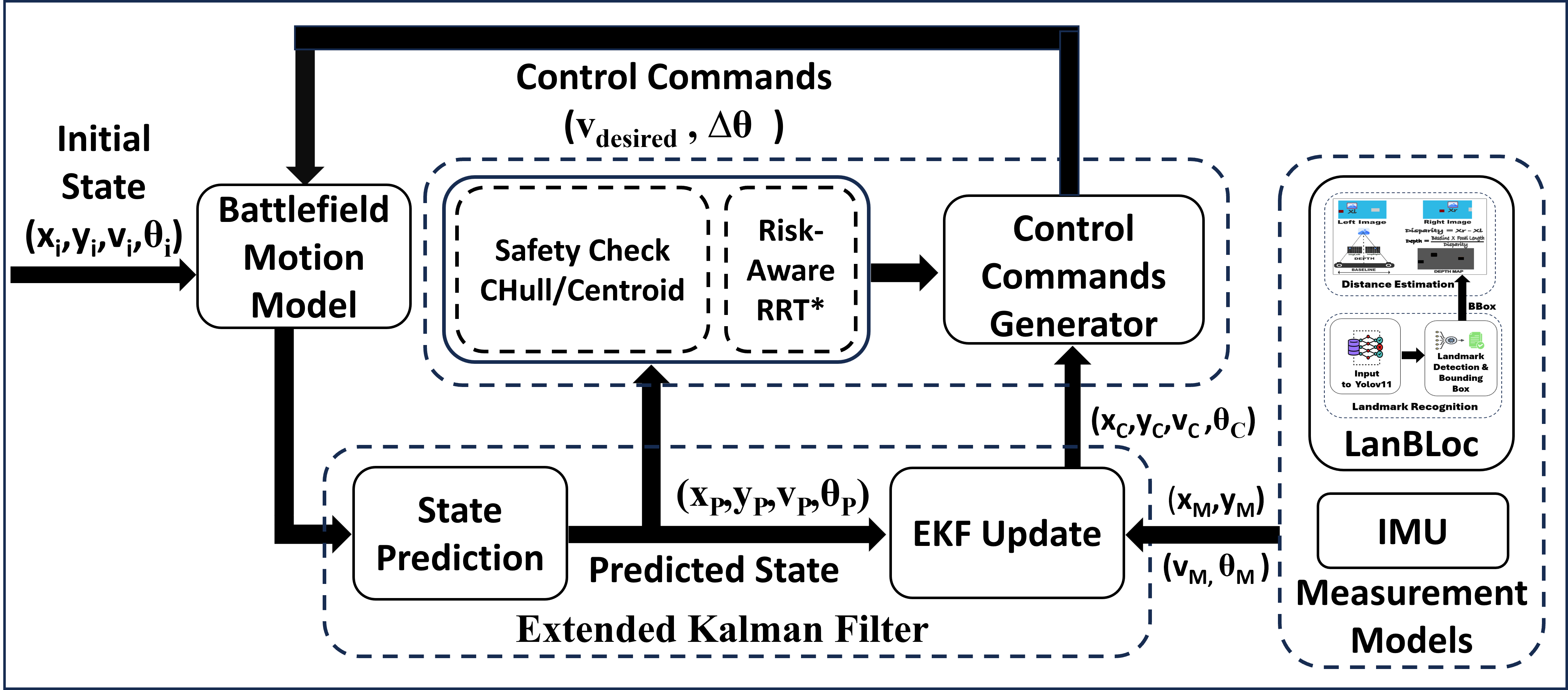}
\caption{An Overview of Safe Path Navigation Framework for GPS-denied Environment using Battlefield Motion Model (BMM), Extended Kalman Filter (EKF), and safety-checking algorithms (CHull and Centroid). } 
\label{syst_overview}
\end{figure}

In the context of battlefield navigation, a safe path refers to a field that avoids obstacles and possible hazards. A continuous function that maps a parameter $s$ representing a sequence to a point in the navigable space, such that the path avoids hazards, and collisions with obstacles and respects the constraints of the navigable space. Mathematically, a safe path \( \mathbf{p}(s) \) can be defined as follows:
\begin{equation}
     \mathbf{p}(s) : [s_0, s_f] \rightarrow \mathbb{R}^n
     % \label{path_representation}
\end{equation}
where $s$ is the parameter, representing sequence, with  $s_0$ and $s_f$ being the start and end sequence numbers, $\mathbb{R}^n$ represents the n-dimensional navigable space, $ \mathbf{p}(s)$ is the position in the navigable space.
The path starts at the initial position $p_0$ and ends at the target position $pf$, which is given by:
\begin{equation}
        p(s_0)=p_0, \quad
        p(s_f)=p_f
\end{equation}
The safe path is generated by the control center based on the current state of the moving node and obstacle/hazard components cluttered in the region of interest. The node initially obtains the safe path as a series of path segments $(s_1,s_2 \ldots s_n)$  each containing trajectory points  $pi=\{(x_{i1},y_{i1}),(x_{i2},y_{i2}),\ldots,(x_{im},y_{im})\}$ from the command and control center via secure communication. For each path segment $s_i = (p_1,p_2,\ldots p_m)$, the node can use either SafeNav-Chull or SafeNav-Controid algorithms to compute control commands. Fig. \ref{syst_overview} shows an overview
of our proposed navigation framework for GPS-denied battlefield environments. The moving entity uses a BMM \& EKF for motion planning, LanBloc \& IMU for state measurement, and  SafeNav-Chull or SafeNav-Controid for heading and maneuvering decisions.
% using a battlefield-specific motion model integrated with an Extended Kalman Filter provided with the LanBLoc algorithm as a location measurement model.
%%%%%%%%%%%%%%%%%%%%%%%%%%%%%%%%%%Explanation of System Overview%%%%%%%%%%%%%%%%%%%%%%%%%
The process begins with the initial State $(x_i,y_i,v_i,{\theta}_i)$ which includes the initial position coordinates $(x_i,y_i)$ , speed $(v_i)$ and heading angle $({\theta}_i)$ of the moving entity. The moving entity runs the stereo vision and YOLOv11-based localization algorithm $(LanBLoc)$ to obtain its initial position and assumes to establish communication with the control center via a secure communication protocol. However, the architecture of node communication with a control center is not covered in this research work. 
% The Battlefield Motion Model simulates the dynamics of the object within the battlefield environment. It takes the initial state  $(x_i,y_i,v_i,{\theta}_i)$ and control commands $(v_{desired},\Delta\theta)$ to predict the next possible state $(x_{P},y_{P},v_{P},{\theta}_P)$. The EKF Update uses the predicted position from the battlefield motion model and the measurements from the LanBLoc $(x_{P},y_{P})$ to correct the predicted state. The EKF compares the predicted state $(x_{P},y_{P},v_{P},{\theta}_P)$ with the measured state $(x{M},y_{M},v_{M},{\theta}_M)$ to generate an updated and corrected state. The corrected state is then fed back into the battlefield motion model, allowing the system to iteratively refine the mobile node’s trajectory over time.
%%%%%%%%%%%%%%%%%%%%%%%%%%%%%%%%System Model%%%%%%%%%%%%%%%%%%%%%%%%%%%%%%%%%%%%%%%%%%%%%

%The system works under the assumption that the command and control server generates and provides the safe path information %shown in Fig. \ref{safe_positions} to the moving entity. 
 % shown in Fig.\ref{lanbloc_architecture} and discussed in detail in Section \ref{subsec_lanbloc_algorithm}.

\section{Proposed Methodology}
This section provides a detailed explanation of the proposed methodologies in the Safe Path Navigation Framework shown in Fig. \ref{syst_overview}.
\subsection{Landmark-based localization (LanBLoc) of moving object}
\label{subsec_lanbloc_algorithm}

%%%%%%%%%%%%%%%%%%%%%%% lanbloc architecture%%%%%%%%%%%%%%%%%%%%%%%%%%%%%%%%%%%%%%%%%%%%%
%\begin{figure}[!ht]
%\includegraphics[width=0.48\textwidth]{system_overview_1.png}
%\centering
%\caption{System Overview of LanBloc framework. Distance estimation and landmark recognition steps and the results after two operations are fused to calculate the position of an unknown node.} 
% \label{lanbloc_architecture}
% \end{figure}
%%%%%%%%%%%%%%%%%%%%%%%%%%%%%%%%%%%%%%%%%%%%%%%%%%%%%%%%%%%%%%%%%%%%%%%%%%%%%%%%%%%%%%%%%
%Deploying anchor nodes in Wireless Sensor Networks (WSNs) can be particularly demanding, especially in harsh environments like battlefields, where network connectivity is often sparse and unstable. In such scenarios, traditional anchor nodes struggle to maintain communication and update critical localization information due to their limited coverage and the dynamic nature of the network. Additionally, the energy and bandwidth costs associated with broadcasting beacon packets from physical anchors are high. 
To address the challenges of deploying physical anchor nodes, recent research presents a solution by recognizing the geographical landmarks as reference points using the YOLOv8-based landmark recognition model\cite{lanbloc_wowmom}. We improve %this model  we present the state-of-the-art 
this landmark recognition model using \text{YOLOv11s} \cite{yolo11v_ultralytics} to identify and precisely locate the reference landmarks. %The YOLOv11s architecture uses the C3k2 block for efficient feature extraction, the SPPF module for better multi-scale feature aggregation, and the C2PSA block for improved spatial attention compared to its predecessors. So, 
We utilized the pre-trained backbone of YOLOv11s and added a custom classifier head for landmark classification and bounding box coordinates. We utilized Generalized Intersection over Union (GIoU) for bounding box loss, Cross-entropy (CE) for landmark classification loss, and Binary cross-entropy (BCE) for the objectness score. The model was trained for $100$ epochs with a batch size of $16$ using $SGD$ optimizer on the real-world landmark dataset (MSTLandmarkv1) \cite{lanbloc_dataset} containing 34 landmark classes. We applied pre-processing and augmentation on the image data before training by flipping, rotating, scaling, noising, and lighting variations to make the landmark dataset more realistic to battlefield scenarios and robust to variations in input data.

The landmark recognition model achieved high performance for training and validation. For training, it demonstrated a Box Loss Precision (Pr) of 0.938, Recall (R) of 0.963, and mAP @ 0.5 IoU of 0.968, with mAP @[0.5:0.95] IoU at 0.841. Validation results were similarly strong, with a Precision of 0.931, Recall of 0.953, mAP @ 0.5 IoU of 0.964, and mAP @[0.5:0.95] IoU at 0.834 across all landmark classes. %These metrics indicate the model's robust accuracy and generalization.

The distances between the moving entity and the identified landmark anchors are calculated using an efficient stereo-matching technique \cite{lanbloc_dataset} with high accuracy. Using the localized landmark anchors and the calculated distances to the anchors, the LanBLoc algorithm computes the position of a moving object with an average RMSE of 0.0142m in the X-position and 0.039m in the Y-position.  By utilizing the recognized landmarks as anchor nodes for trilateration, we enhanced the accuracy and reliability of node localization with further position optimization. % as shown in Fig.\ref{lanbloc_architecture}. %Unlike traditional radio-based anchor nodes, physical landmarks are robust and enduring, making them ideal for challenging environments. Furthermore, this approach extends the applicability of our framework to sparse sensor networks deployed across extensive geographical areas. In essence, this approach offers a more practical and resilient approach to anchor node deployment and localization, particularly in rugged settings like battlefields, where deploying and maintaining physical anchors are impractical as well as dangerous.

\subsection{Motion Planning of Moving Entity in the Battlefield}
%%%%%%%%%%%%%%%%%%%%%%%%%%%%%%%%%%%%%%%%%%%%%%%%%%%%%%%%%%%%%%%%%%%%%%%%%%%%%%%%%%%%%%%%%
\label{subsec_motion_model}
Developing a motion planning of a moving entity for a battlefield scenario involves considering both the physical aspects of motion and the tactical constraints imposed by the battlefield. Our \textbf{Battlefield Motion Model (BMM)} incorporates kinematic aspects of the vehicle or troop unit such as velocity, acceleration, and deceleration limits, and its maneuverability, which indicates the maximum rate of change of heading. Given a control command and current state, a predicting function updates the unit's next state based on those factors. The BMM is represented in Eqns: \ref{vel_update},\ref{head_update}, and \ref{pos_update}. 
%This model incorporates kinematic aspects of a moving object such as position, velocity, acceleration, and heading, as well as dynamic control inputs (commands) for desired speed and heading change. 
The parameters are given in the table \ref{tab:parameters}. 
%%%%%%%%%%%%%%%%%%%%%%%%%%%%%%%%%%%%%parameters Table%%%%%%%%%%%%%%%%%%%%%%%%%%%%%%%%%%%%
\begin{table}[h!]
\centering
\caption{Parameters used in the Battlefield Motion Model}
\begin{tabular}{cl}
\hline
\textbf{Parameters} & \textbf{Description} \\
\hline
$\vec{p}_t = [x_t, y_t]$ & Position vector at time $t$ \\
$v_t$ & Velocity at time $t$ \\
$\theta_t$ & Heading (direction) at time $t$ \\
$v_{\text{desired}}$ & Desired velocity (speed command) \\
$\Delta\theta$ & Change in heading (heading command) \\
$a$ & Acceleration capability of the unit \\
$d$ & Deceleration capability of the unit \\
$m$ & Maneuverability (maximum rate of change of heading) \\
$\Delta t$ & Time step \\
\hline
\end{tabular}
\label{tab:parameters}
\end{table}
%%%%%%%%%%%%%%%Motion model theory%%%%%%%%%%%%%%%%%%%%%%%%%%%%%%%%%%%%%%%%%%%%%%%%%%%%%%%
Now, the motion model can then be expressed as:
\subsubsection{Velocity Update:}
\begin{equation}
\label{vel_update}
 v_{t+1} = 
   \begin{cases} 
   \min(v_t + a\Delta t, v_{\text{desired}}, v_{\text{max}}) & \text{if } v_{\text{desired}} > v_t \\
   \max(v_t - d\Delta t, v_{\text{desired}}, 0) & \text{if } v_{\text{desired}} \leq v_t 
   \end{cases}
\end{equation}
where, \( v_{\text{max}} \) is the maximum speed of the unit.

\subsubsection{Heading Update:}
\begin{equation}
\label{head_update}
\theta_{t+1} = \theta_t + \text{clip}(\Delta\theta, -m\Delta t, m\Delta t)
\end{equation}
    % The function \( \text{clip}(x, a, b) \) restricts \( x \) to the range \( [a, b] \).
\subsubsection{Position Update:}
% \begin{equation}
% \label{pos_update}
%  \vec{p}_{t+1} = \vec{p}_t + v_{t+1}\cdot \Delta t \cdot \begin{bmatrix} \cos(\theta_{t+1}) \\ \sin(\theta_{t+1}) \end{bmatrix}
% \end{equation}
% \begin{equation}
%     \label{pos_update}
%     p_{x, t+1} = p_{x, t} + v_{t+1} \cdot \Delta t \cdot \cos(\theta_{t+1}), \quad \\
%     p_{y, t+1} = p_{y, t} + v_{t+1} \cdot \Delta t \cdot \sin(\theta_{t+1})
% \end{equation}
% \begin{align}
% \label{pos_update_x}
% p_{x, t+1} &= p_{x, t} + v_{t+1} \cdot \Delta t \cdot \cos(\theta_{t+1}), \\
% \label{pos_update_y}
% p_{y, t+1} &= p_{y, t} + v_{t+1} \cdot \Delta t \cdot \sin(\theta_{t+1}).
% \end{align}

\begin{equation}
\label{pos_update}
 \vec{p}_{t} = \vec{p}_{t-1} + v_{t-1}\cdot \Delta t \cdot \begin{bmatrix} \cos(\theta_{t-1}) \\ \sin(\theta_{t-1}) \end{bmatrix}
\end{equation}
In this model, the velocity at the next time step, $v_{t+1}$, is updated based on the current velocity, $v_t$  constrained by the acceleration/deceleration capabilities and speed limits. The heading update adjusts the heading, $\theta_{t+1}$, by a controlled amount, $\Delta\theta$, within maneuverability limits. The position update calculates the new position, $\vec{p}_{t+1}$, based on the updated velocity and heading. This model assumes a linear motion and instant change of speed and heading for simplicity.

However, in practice, the dynamics of the battlefield are unpredictable due to changing environmental conditions, terrain, and obstacle components. To represent the motion model of the real-battlefield scenario, we need to consider the process noise in position, velocity, acceleration, and heading that capture the uncertainty and variability due to factors like terrain and other unmodeled influences such as obstacles and hazards. It includes terms for terrain effects $(\tau)$ and process noises ${\epsilon}_{x,t}, {\epsilon}_{y,t},{\epsilon}_{v_x,t},{\epsilon}_{v_y,t},{\epsilon}_{a_x,t},{\epsilon}_{a_y,t},{\epsilon}_{\theta,t}$ in as position, velocity, acceleration and heading respectively. Hence, the motion model is defined in Eqns. \ref{vel_update},\ref{head_update}, and \ref{pos_update} is updated as:
% \begin{equation}
% \label{vel_update2}
%     v_{t+1} = f(v_t, a_t, \epsilon_{v,t},{\epsilon}_{a,t}) 
% \end{equation}
% %%%%%%%%%%%%%%%%%%%%%%%%%%%%%%%%%%%%%%%%%Heading Update%%%%%%%%%%%%%%%%%%%%%%%%%%%%%%%%%%
% \begin{equation}
% \label{head_update2}
%      \theta_{t+1} = \theta_t + \Delta \theta + \epsilon_{\theta,t} 
% \end{equation}
\begin{equation}
v_{t+1} = f(v_t, a_t, \epsilon_{v,t}, \epsilon_{a,t}), \quad 
\theta_{t+1} = \theta_t + \Delta \theta + \epsilon_{\theta,t}
\label{combined_update}
\end{equation}

where $\Delta \theta=\arctan({\Delta x / \Delta y})$ rate of heading change, and $f$ is a function that computes the next velocity based on the current velocity, acceleration, and terrain. Next, the position of ME is estimated according to Eqn. \ref{pos_update2}
%%%%%%%%%%%%%%%%%%%%%%%%%%%%%%%%%%%%%%%%%Position Update%%%%%%%%%%%%%%%%%%%%%%%%%%%%%%%%%%
\begin{equation}\label{pos_update2}
    \begin{split}
        x_{t+1} = x_t + v_{t+1} \cdot \Delta t \cdot \cos(\theta_{t+1})  + \epsilon_{x,t}
        \\y_{t+1} = y_t + v_{t+1} \cdot \Delta t \cdot \sin(\theta_{t+1}) + \epsilon_{y,t}
    \end{split}
\end{equation}
%%%%%%%%%%%%%%%%%%%%%%%%%%%%%%%%%%%Terrian Effect Theory%%%%%%%%%%%%%%%%%%%%%%%%%%%%%%%%%%
Terrain effect $\tau$ can significantly influence the error in the predicted position. By accurately computing $\tau$, predictions for the ME's position can be made more precise, enhancing the navigation strategy in challenging and dynamic terrain conditions.
While moving from point P1 \( (x1, y1) \) to P2 \( (x2, y2) \), 
%calculated distance $d_c$ and 
total time to travel $t_{\tau}$ considering $\tau$ is given by Eqn \ref{combined_time_distance}.
% \ref{dist_travelled} and \ref{time_tau}
   % \begin{equation}
   % \label{dist_travelled}
   %     d_c = \left |{\left |{{P_1 - P_2}}\right |}\right |
   % \end{equation}

   % \begin{equation}
   % \label{time_tau}
   %       t_{\tau} = \frac{d_c}{v_c} + \lambda 
   % \end{equation}
  \begin{equation}
t_{\tau} = \frac{\left\| P_1 - P_2 \right\|}{v_c} + \lambda
\label{combined_time_distance}
\end{equation}

 where $\lambda = \tau / v_c $ is the time delay due to $\tau$. $\lambda$ is calculated at the beginning with prior knowledge of  $\tau$. However, $\tau$ may change as navigating along the field due to changing terrain conditions and the velocity. Thus, the predicted position $p_{t+1}$ of moving entity considering the terrain effect is given by Eqn. \ref{predicted_pos}.
%%%%%%%%%%%%%%%%%%%%%State Updates with Terrian Effect %%%%%%%%%%%%%%%%%%%%%%%%%%%%%%%%%
\begin{equation}
\label{predicted_pos}
    % Predicted ~ Position: = Xc + (Vt \cdot t \cdot cos(\theta i), Vt \cdot t \cdot sin(\theta i))
     \\p_{t+1} = p_t + v_{t+1}\cdot t_{\tau} \cdot \begin{bmatrix} \cos(\theta_{t+1}) \\ \sin(\theta_{t+1}) \end{bmatrix} + \begin{bmatrix} \epsilon_{x,t}\\ \epsilon_{y,t} \end{bmatrix}
\end{equation}
where ${p}_t$ is the current position of moving entity. 
%%%%%%%%%%%%%%%%%%%%%%%%%%%%%%%%%%%%%%%%%%%%%%%%%%%%%%%%%%%%%%%%%%%%%%%%%%%%%%%%%%%%%%%%%%%%%
% Now, the position error is given by Eqn. \ref{pos_err}
%    \begin{equation}
%    \label{pos_err}
%        % e = \Delta d =  \sqrt{(x_p-x_m)^2+(y_p-y_m)^2}
%        e = \Delta d =  \left |{\left |{{p_{t+1} - P_m}}\right |}\right |
%    \end{equation}

%    where, ${p}_{t+1} = (x_p,y_p)$ is predicted position and $p_m = (x_m, y_m)$ is measured position. So, the difference in the position is 
%    $\Delta d$ such that \\\\
%    $\Delta t = \frac{\Delta d}{v_c} $
%    which is equivalent to 
%    \begin{equation}
%         \lambda = \frac{ \tau }{v_c}
%    \end{equation}
% The error in the predicted distance compared to the measured distance is caused by the unusual terrain effect T. So we can utilize this $\tau$ value to compute the travel time and next possible position more precisely using Eqn. \ref{time_tau} and \ref{predicted_pos} respectively.
%%%%%%%%%%%%%%%%%%%%%%%%%%%%%%%%%%%%%%%%%%%%%%%%%%%%%%%%%%%%%%%%%%%%%%%%%%%%%%%%%%%%%%%%%%%%
The Eqn. \ref{predicted_pos} captures the essential dynamics of the motion model for any moving entities %for eg: a ground vehicle or troop unit 
on a battlefield, considering their physical capabilities and the control inputs for navigation.

\subsection{Trajectory Estimation using BMM and EKF}
\label{subsec_ekf_prediction}
%On the battlefield, state tracking and navigation systems are often non-linear due to complex terrain, dynamic obstacles, and hazardous components that should be avoided by moving entities to ensure their survivability.
The Extended Kalman Filter (EKF) for the non-linear battlefield scenario %is an extension of the Kalman filter for nonlinear systems. It 
linearizes the current mean and covariance, allowing it to approximate the state of a nonlinear system, 
and thus is an effective filtering algorithm due to its low computational complexity and high estimation accuracy. 
We adapted an EKF by integrating BMM defined in Eqn. \ref{combined_update} and \ref{pos_update2} in its prediction step to predict the future state of a moving entity. The mathematical representation of EKF is shown below:
\begin{algorithm}
    \caption{Safety Check using Convex Hull Method }
    \label{alg_ConvexHull}
    \begin{algorithmic}[1]
    \Require current position $p_t$, $v_t$, $\theta_t$, $a_t$, $\epsilon_{v,t}$, $\epsilon_{\theta,t}$, $t_{\tau}$, list of path segments $S = \{s_i..s_n\}$
    \Ensure Estimated Position $p_{t+1}$ and Control Input U
    \Procedure{ConvexHull}{$p_t$, $v_t$, $\theta_t$, $a_t$, $\epsilon_{v,t}$, $\epsilon_{\theta,t}$, $t_{\tau}$}
        \State Update $v_{t+1} \leftarrow f(v_t, a_t, \epsilon_{v,t})$
        \State Update $\theta_{t+1} \leftarrow \theta_t + \Delta \theta + \epsilon_{\theta,t}$
        \State Estimate: $
p_{t+1} \leftarrow p_t + \big(v_{t+1} \cdot t_{\tau} \cdot \cos(\theta_{t+1}), v_{t+1} \cdot t_{\tau} \cdot \sin(\theta_{t+1})\big)$
% $n$ segments $s_{i-n}\in S$
     \ForEach{$segment$ $s_{i}$ in $S$}
        \State $current\_hull \leftarrow ConvexHull(segment)$
        \State $next\_segment\_index \leftarrow$ index of next segment
        \If{$next\_segment\_index$ exists}
            \State $next\_hull \leftarrow ConvexHull(next\_segment)$
        % \Else
        %     \State $next\_hull \leftarrow$ None
        \EndIf
        \If{PointInHull($predicted\_position$, $current\_hull$)
            \hspace{0cm} or ($next\_hull$ is not None \\
            \hspace{1.5cm})}
            \State \Return $p_{t+1}$
        \Else
            \State Compute: ControlInput($current\_hull$,$next\_hull$)
            \State break
        \EndIf
    % \EndFor
     \State \Return  $p_{t+1}$ and U($v_i$,${\theta}_i$)
    \EndProcedure
    \end{algorithmic}   
\end{algorithm}
\subsubsection{State and Measurement Models}
The state of the system at time step $k$ is represented by the vector $x_k$. 
%The EKF models the system dynamics and measurement with the following nonlinear functions:
The state transition function $X_{k+1}$ and the measurement function $Z_{k+1}$ are represented in Eqn. \ref{state_transition_func}:
    % \begin{equation}
    % \label{state_transition_func}
    % \begin{split}
    %     X_{k+1} = f(X_{k},U_{k})+ W_{k}\\
    %     Z_{k+1} = h(X_{k+1})+V_{k+1}
    % \end{split}
    % \end{equation}
\begin{equation}
X_{k+1} = f(X_{k},U_{k})+ W_{k}, \quad
Z_{k+1} = h(X_{k+1})+V_{k+1}
\label{state_transition_func}
\end{equation}

$f(\cdot)$ is the nonlinear state transition function defined in Eqn \ref{predicted_pos}, which predicts the next state based on control input $U_k$ and the previous state $X_k$ while the $h(\cdot)$ is the LanBloc measurement function, that maps the true state space into the observed space.$W_k$ is the process noise and and $V_{k+1}$ is the measurement noise which is normally distributed with covariance $Q_k$ and $R_{k+1}$ respectively. %The battlefield navigation adopts the \textbf{LanBLoc} as a measurement function.
\subsubsection{Prediction Step}
The EKF predicts the state and error covariance at the time interval $k$. The predicted state estimate $\hat{X}_{k+1}$ and predicted covariance estimate $P_{k+1}$ are given by: 
% \begin{equation}
% \label{predicted_state}
% \begin{split}
%       \hat{x}_{k+1}=f(\hat{x}_k,U_k) + w_{k}\\
%      P_{k+1}= F_k \cdot P_k \cdot F_{k}^T+ Q_k
% \end{split}
% \end{equation}
\begin{equation}
\label{predicted_state}
    \hat{x}_{k+1}=f(\hat{x}_k,U_k) + w_{k}, \quad
     P_{k+1}= F_k \cdot P_k \cdot F_{k}^T+ Q_k
\end{equation}

where $F_k$ is the Jacobian of state transition function $f(\cdot)$ with respect to $X$ that is evaluated at $\hat{X}_{k+1}$ and $Q_k$ is the covariance matrix of process noise. %The Jacobian is the matrix of all partial derivatives of the vector of $f(\cdot)$ and $h(\cdot)$ around the estimated state.
\subsubsection{Update Step}
The EKF incorporates the location measured by the LanBloc algorithm to update the state estimate and covariance. The measurement residual $\Tilde{y}_{k+1}$ and residual covariance $S_{k+1}$ are calculated using Eqn. \ref{residual_cov}.
% \begin{equation}
% \label{residual_cov}
% \begin{split}
%      \Tilde{y}_{k+1} = z_k-h(\hat{x}_{k+1})\\
%     S_{k+1}=H_{k+1}\cdot P_{k+1}\cdot H_{k+1}^T+R_{k+1}
% \end{split}
% \end{equation}
\begin{equation}
\label{residual_cov}
     \Tilde{y}_{k+1} = z_k-h(\hat{x}_{k+1}),\quad
    S_{k+1}=H_{k+1}\cdot P_{k+1}\cdot H_{k+1}^T+R_{k+1}
\end{equation}
Where $z_{k+1}$ represents the actual measurement of the position at time step $k+1$.
The Kalman Gain is a fundamental component of the Kalman Filter that determines the degree to which the new measurement will be incorporated into the state estimate. 
% The calculation of the Kalman Gain balances the uncertainty in the current state estimate $($as represented by the state covariance matrix$)$ and the uncertainty in the new measurement as represented by the measurement noise covariance.
So, the Kalman Gain $ K_{k+1}$ is computed using the predicted covariance estimate $P_{k+1}$, the transpose of the observation model $H_{k+1}^T$, and the inverse of the residual covariance $S_{k+1}^{-1}$ in equation \ref{kalman_gain} below.
\begin{equation}
\label{kalman_gain}
    K_{k+1}=P_{k+1} \cdot H_{k+1}^T \cdot S_{k+1}^{-1}
\end{equation}
Finally, the state estimate $\hat{x}_{k+1}$ and covariance estimate $P_{k+1}$ are updated in Eqn. \ref{state_update}.
% \begin{equation}
%     \label{state_update}
%     \begin{split}
%      \hat{x}_{k+1} = \hat{x}_k +K_{k+1}\cdot \Tilde{y}_{k+1}\\
%         P_{k+1} = (I-K_{k+1}\cdot H_{k+1}) \cdot P_{k+1}
% \end{split}
% \end{equation}
\begin{equation}
    \label{state_update}
     \hat{x}_{k+1} = \hat{x}_k +K_{k+1}\cdot \Tilde{y}_{k+1}, \quad
        P_{k+1} = (I-K_{k+1}\cdot H_{k+1}) \cdot P_{k+1}
\end{equation}

Where $I$ is the identity matrix.

\subsection{Safety Check and Controlled Maneuver}
\subsubsection{Convex Hull Method}
\label{subsec_safety_check}
% The convex hull $C$ of points  $p_1, ..., p_N$ is defined by \cite{convex_hull_app2} as, 
% \begin{equation}
% \label{convex_hull_eq}
%     C = \left\{ \sum_{j=1}^{N} \lambda_j p_j \mid \lambda_j \geq 0, \sum_{j=1}^{N} \lambda_j = 1 \right\} 
% \end{equation}
% where $\lambda$ represents a set of coefficients in the linear combination of points that form the convex hull.
% Chan’s algorithm \cite{chans_algo} %is shown in \textbf{Algorithm \ref{algo_chan}}
% computes the convex hull with the time complexity of $O(n \log h)$, where $n$ indicates the total number of points in a cluster and $h$ indicates the number of vertices of the convex hull polygon. The algorithm divides the set of given points into smaller subsets, finds the convex hull on each subset, and then merges those hulls. The merging phase utilizes the idea of the tangent finding process similar to the Graham scan algorithm. %shown in \textbf{Algorithm \ref{algo_ghram_scan}}.
% \vspace{-0.1in}

The convex hull technique provides a geometrically sound and computationally efficient method to determine the boundaries of a safe path, making it an essential tool in navigation and path-planning applications. Our method utilizes the convex hull approach\cite{convex_hull_app2} to confirm the safe zone while making the heading decision toward the look-ahead point while navigating. %using Chan’s algorithm \cite{chans_algo} to compute the convex hull of the current and consecutive segment with the time complexity of $O(n \log h)$.
For safety checks, the path is split into multiple segments and arranged according to their sequence. For each segment, $S =$\{($x_1$,$y_1$),($x_2$,$y_2$),...,($x_n$,$y_n$)\}, convex hull is calculated using Chan’s algorithm \cite{chans_algo} and the predicted state is checked against the convex hull boundary exploiting the point in polygon test \cite{even_odd_algo}.

At the beginning of each trial, the node obtains its initial position  $(x_{i},y_{i}$ using LanBLoc while velocity $(v_i)$ and heading $(\theta_i)$ is measured using IMU sensor simulator which serves as an input to EKF module that predicts the next state $(x_{P},y_{P},v_{P},{\theta}_P)$ based on the motion model discussed in detail in Section \ref{subsec_motion_model}. Then, the node checks if the predicted position lies within the convex hull indicating the safe zone, and makes the heading decisions. If the position lies within the convex hull of the current or next segment, it moves with the default command (desired speed, desired heading). Otherwise, it re-calculates the control commands. Node steps back with the default control command if the predicted position is not within any of the adjacent hulls(condition when a node is in the boundary of the safe path.)
The trajectory prediction mechanism is explained in detail in Section \ref{subsec_ekf_prediction} while the safety check using a convex hull is discussed in Algorithm \ref{alg_ConvexHull}.

\subsubsection{Centroid Method}

The Centroid Method relies on the centroids of sub-segments within a safe path to compute the control commands. In this approach, the current path segment is divided into smaller sub-segments, and the centroids of these sub-segments are calculated. The algorithm determines the closest centroid to the entity's current position(updated by EKF) and computes control input $U(v_i,{\theta_i})$ required to reach the centroid. Using these control commands, it predicts the next position of the entity which is mentioned in detail in the Algorithm \ref{alg_CentroidMethod}. This algorithm attracts the moving entity toward the centroid of the path segment with an objective of minimizing the drift from the central trajectory.
% If the predicted position lies within the bounds of the sub-segment, the entity moves toward the centroid. If not, the control inputs are adjusted to guide the entity back within the safe zone. This method ensures precise, controlled navigation while minimizing deviation from the predefined safe path, making it effective for dynamic environments like battlefields.
\begin{algorithm}
    \caption{Safety Check using Centroid Method }
    \label{alg_CentroidMethod}
    \begin{algorithmic}[1]
    \Require current position $p_t$, $v_t$, $\theta_t$, $a_t$, $\epsilon_{v,t}$, $\epsilon_{\theta,t}$, $t_{\tau}$, list of path segments $S = \{s_i..s_n\}$
    \Ensure Estimated Position $p_{t+1}$ and Control Input U
    \Procedure{EstimateNextPosition} {$p_t$, $v_t$, $\theta_t$, $a_t$, $\epsilon_{v,t}$, $\epsilon_{\theta,t}$, $t_{\tau},s_i$}
        \ForEach{$segment$ $s_{i}$ in $S$}
            % \State For $n$ segments $s_{i-n}\in S$ 
            \State calculate centroids $C \leftarrow \{c_j..c_n\}$
        % with centroid $(x_j, y_j)$ in the list of path segments}
            \State Select $c_j \leftarrow  \arg\min_{c_j \in C} \|c - p_c\|$
            % \State Calculate distance to centroid: $d_c \leftarrow \sqrt{(x_i - x_j)^2 + (y_i - y_j)^2}$
            \State Compute U($v$,${\theta}_i$): $v_i = v_t$,$\Delta \theta=\arctan({\Delta x / \Delta y})$ 
            \State Update $v_{t+1} = f(v_t, a_t, \epsilon_{v,t}) $, $\theta_{t+1} \leftarrow \theta_t + \Delta \theta + \epsilon_{\theta,t}$ 
            \State $p_{t+1} \leftarrow p_t  + (v_{t+1} * t_{\tau} * \cos(\theta_{t+1}), v_{t+1} * t_{\tau} * \sin(\theta_{t+1}))$
            % \If{$p_{t+1}$ $\in s_i$}
            % \State \Return $p_{t+1}$
            % \Else
            %     % \State Stop moving to the predicted position
            %     \State Select $c_k \leftarrow \arg \min_c \in {C_i - c_j} \|c - P_f\|$
            %     % \State 
            %     \State Continue $ 1 - 8 $ to adjust control input $U_i$ for the next target position $c_k$
            % \EndIf
            \State \Return  $p_{t+1}$ and U($v_i$,${\theta}_i$)
    % \EndFor
     \EndProcedure
    \end{algorithmic}
\end{algorithm}

\subsection{Risk minimization and Obstacle avoidance using Risk-Aware RRT*}
\subsubsection{Rapidly Exploring Random Tree-Star (RRT*)}
RRT* \cite{rrt_star1} relies only on the path length to select the optimal trajectory. During the optimization phase RRT* searches to find the shortest path between consecutive nodes $\mathbf{x}_i$ and $\mathbf{x}_{i+1}$ using the cost function in Eqn \ref{rrt-cost}.

\begin{equation}
    Cost(p) = \sum_{i=1}^{n-1} \|\mathbf{x}_i - \mathbf{x}_{i+1}\|
    \label{rrt-cost}
\end{equation}

\subsubsection{Risk-Aware RRT*} We propose a new variant of RRT* called Risk-Aware RRT* (RAw-RRT*) modifying its cost function in Eqn. \ref{rrt-cost} which ensures the obstacle-free optimal path with the shortest length and minimal risk of hazards. We use this algorithm for re-generating the path that avoids obstacles along the predicted next position while running CHull and Centroid-based Controlled maneuvering. For a trajectory with points $\{\mathbf{p}_1, \mathbf{p}_2, \dots, \mathbf{p}_n\}$, the Risk-Aware cost function in RAw-RRT* is defined as:
\begin{equation}
   C_{\text{total}} = \alpha \cdot C_{\text{length}} + \beta \cdot C_{\text{risk}}
\end{equation}

% \begin{equation}
%     C_{\text{length}} = \sum_{i=1}^{n-1} \| \mathbf{p}_i - \mathbf{p}_{i+1} \|
% \end{equation}

% \begin{equation}
%     C_{\text{risk}} = \sum_{i=1}^{n} \text{WRS}(\mathbf{p}_i)
% \end{equation}

Where, $C_{\text{total}}$ is the Total cost of a trajectory, $C_{\text{length}}$ is the total path length calculated as the sum of Euclidean distances between consecutive points. $C_{\text{risk}}$ is the total risk cost, calculated as the sum of Weighted Risk Scores (WRS) for each point $\mathbf{p}_i$ in the trajectory. $\alpha, \beta$ are the weight parameters that control the trade-off between path length and risk.

\subsubsection{Edge-Wise WRS Calculation}
For each edge in the trajectory, the Weighted Risk Score (WRS) is calculated based on the deviation of the edge from the central trajectory. Let $\mathbf{c}_i$ represent the central trajectory point closest to $\mathbf{p}_i$, then:

\text{A. Deviation:}
\begin{equation}
    d_i = \| \mathbf{p}_i - \mathbf{c}_i \|  
\end{equation}

\text{B. Zone-Based Weighted Risk Score :}
% \begin{equation}
%     \text{WRS}(\mathbf{p}_i) =
% \begin{cases} 
% \frac{d_i}{w_1}, & \text{if } d_i \leq 2.0 \text{ (Zone 1)} \\
% \frac{d_i}{w_2}, & \text{if } 2.0 < d_i \leq 4.0 \text{ (Zone 2)} \\
% \frac{d_i}{w_3}, & \text{if } 4.0 < d_i \leq 6.0 \text{ (Zone 3)} \\
% \frac{d_i}{w_4}, & \text{if } 6.0 < d_i \leq 8.0 \text{ (Zone 4)} \\
% 8.0, & \text{if } d_i > 8.0 \text{ (Beyond buffer)}
% \end{cases}
% \end{equation}

% Where, $w_1, w_2, w_3, w_4$ represent the risk weights assigned to each zone.

\begin{equation}
\label{wrs}
  \text{WRS}(\mathbf{p}_i) = \begin{cases} 
|d_i| \times \left( \frac{|d_i|}{w_j} \right) & \text{if } a_j \leq |d_i| < b_j\\ 
% \quad \text{for} \ j = 1, 2, 3, 4,...N \\
d_{\text{max}} \times \left( \frac{d_{\text{max}}}{w_{\text{max}}} \right) & \text{if } |d_i| \geq d_{\text{max}} \\
% \quad (d_{\text{max}} = 8, w_{\text{max}} = \text{maximum weight})
\end{cases}
\end{equation}
Where,
$|d_i|$ is the absolute deviation for point $\mathbf{p}_i$.
 $a_j$  and $b_j$  are the lower and upper bounds of zone $Z_j$ and $w_j$ is the maximum risk weight of $Z_j$ .
 $d_{\text{max}}$  is the maximum deviation and
$w_{\text{max}}$ is the maximum weight assigned when $|d_i| \geq d_{\text{max}}$.
For $|d_i| < d_{\text{max}}$, the Eqn. \ref{wrs} computes the weighted risk score based on the absolute deviation and the zone-specific weight $w_j$.

For $|d_i| \geq d_{\text{max}}$, the deviation is capped at \( d_{\text{max}} \) and a maximum weight  $w_{\text{max}}$ is used to calculate the risk score.

This ensures that both small and large deviations are treated appropriately, with large deviations capped at $d_{\text{max}}$ and assigned a constant maximum risk score.

\subsubsection{Total Cost for Trajectory}
For a trajectory with points $\{\mathbf{p}_i..\mathbf{p_n}\}$, the total risk-aware cost is given by Eqn. \ref{risk-aware-cost}.
\begin{equation}
    C_{\text{total}} = \alpha \cdot \sum_{i=1}^{n-1} \| \mathbf{p}_i - \mathbf{p}_{i+1} \| + \beta \cdot \sum_{i=1}^{n} \text{WRS}(\mathbf{p}_i)
    \label{risk-aware-cost}
\end{equation}
The detailed steps to obtain optimal path using risk-aware cost function are systematically presented in \textbf{Algorithm \ref{alg:RiskAwareRRTstar}}.
%%%%%%%%%%%%%%%%%%%%%%%%%%%%%%%%%Risk-Aware RRT* Algorithm%%%%%%%%%%%%%%%%%%%%%%%%%%%%%%%%
\begin{algorithm}[h]
\caption{Risk-Aware RRT*}
\label{alg:RiskAwareRRTstar}
\begin{algorithmic}[1]

\Require Start position $p_{\text{start}}$, goal position $p_{\text{goal}}$, obstacle map $O$, risk weights $w_i$
\Ensure Optimal path $P_{\text{optimal}}$

\State Initialize tree $T \leftarrow \{p_{\text{start}}\}$, set cost $C(p_{\text{start}}) \leftarrow 0$

\For{$k = 1$ to $N_{\text{max}}$}
    \State Sample $p_{\text{rand}}$, find $p_{\text{nearest}} \in T$, and steer to $p_{\text{new}}$
    \If{$\text{CollisionFree}(p_{\text{nearest}}, p_{\text{new}}, O)$}
        \State Compute: $
        C_{\text{total}}(p_{\text{new}}) = C_{\text{total}}(p_{\text{nearest}}) + \|p_{\text{nearest}} - p_{\text{new}}\| + \text{WRS}(p_{\text{new}})$
        
        \State Add $p_{\text{new}}$ to $T$ and rewire nearby nodes for cost optimization
    \EndIf
\EndFor

\State Extract optimal path $P_{\text{optimal}}$ from $T$ 
\State \Return $P_{\text{optimal}}$

\end{algorithmic}
\end{algorithm}

%%%%%%%%%%%%%%%%%%%%%%%%%%%%%%%%%Safe Navigation Algorithm%%%%%%%%%%%%%%%%%%%%%%%%%%%%%%%%
\subsection{Safe Path Navigation using SafeNav-Chull and SafeNav-Centroid and RAw-RRT*}
The complete steps to guide the object from a start to the end position along a pre-defined safe path are presented in Algorithm \ref{alg:safeNavigation}. This algorithm includes the navigation steps for both SafeNav-Chull and SafeNav-Centroid when integrated with RAw-RRT*.
\begin{algorithm}
\caption{Safe Path Navigation}
\label{alg:safeNavigation}
\begin{algorithmic}[1]
\Require Initial position $p_t$, $v_t$, $\theta_t$, $a_t$, $\epsilon_{v,t}$, $\epsilon_{\theta,t}$, $t_{\tau}$, control input ${U_i}$, path segments $S = \{s_i..s_n\}$  %with centroids $[(x_1, y_1), (x_2, y_2), \ldots, (x_n, y_n)]$
\Ensure Predicted and Actual trajectory list

% \State Initialize trajectory container: $predicted \leftarrow \{(x_i, y_i)\}$ and $measured \leftarrow \{\}$
% \State Set initial control input $U_i = (v_i, \theta_i)$
% \State Sort the path segments in the order of their proximity to the final destination $P_f = (x_f,y_f)$.
\State Select the Method $\in$ \{\textbf{Algorithm} \ref{alg_ConvexHull}, \textbf{Algorithm} \ref{alg_CentroidMethod}\}
\For{$path\_segment$ $\{s_i\} \in S$}
   % \If{method = 'Centroid'}
    \State nextPosition = EstimateNextPosition($p_t$, $v_t$, $\theta_t$, $a_t$, $\epsilon_{v,t}$,$\epsilon_{\theta,t}$,$t_{\tau},s_i$)
    \State $U_i = (v_i, \theta_i)$
    \State Check for Obstacles:
    \If{$ObstacleDetected(p_t, s_i, O)$}
        % \State \textbf{Replan using Risk-Aware RRT*:}
        \State Run \textbf{Algorithm \ref{alg:RiskAwareRRTstar}} to find alternative path to nextPosition that minimize cost.
    \EndIf
    \State Move the object to the nextPosition $p_{t+1}$.
    
    % \ElsIf{method = 'Convex Hull'}
    %     \State nextPosition =  ConvexHull($p_t$, $v_t$, $\theta_t$, $a_t$, $\epsilon_{v,t}$, $\epsilon_{\theta,t}$, $t_{\tau}$) 
    %     \State Check for Obstacles:
    %         \If{$ObstacleDetected(p_t, s_i, O)$}
    %             % \State \textbf{Replan using Risk-Aware RRT*:}
    %             \State Run \textbf{Algorithm \ref{alg:RiskAwareRRTstar}} to find alternative path to nextPosition that minimize cost.
    %         \EndIf
    %     \State Move the object to the nextPosition $p_{t+1}$.
    \State Measure current state using $LanBloc() + IMU + EKF$ 
    % \State Calculate position error $e \leftarrow  \left |{\left |{{p_{t+1} - P_m}}\right |}\right |$
    % \State Update $\lambda \leftarrow e/v_c$
    \State Append predicted state to $predicted$ and measured state to $measured$
    \State Update $U_i$ based on the current and next states 
    % \State \Return $predicted$ and $measured$
    % \EndIf
\EndFor
% \State Repeat until the object reaches the final destination $P_f$. 
\State \Return $predicted$ and $measured$
\end{algorithmic}
\end{algorithm} 
%%%%%%%%%%%%%%%%%%%%%%%%%%%%%%%%%%%%%%%%%%%%%%%%%%%%%%%%%%%%%%%%%%%%%%%%%%%%%%%
%%%%%%%%%%%%%%%%%%%%%%%%%%%%%%%%Experiments and Results%%%%%%%%%%%%%%%%%%%%%%%%
\section{Experiments and Results}
To test the validity of our proposed navigation algorithms, experiments were performed on Alienware Aurora R12 System with 11th Gen Intel Core i7 CPU, 32 GiB Memory, and Nvidia GeForce RTX 3070 GPU using Python 3.11 on PyCharm CE 2024 IDE. The landmark recognition model was trained using Python-3.10.12 and torch-2.5.1+cu121, utilizing NVIDIA A100-SXM4 GPU with 40GB memory.
\subsection{Datasets}
\subsubsection{MSTlandmarkv1 Dataset}
We have used publicly available MSTlandmarkv1\cite{lanbloc_dataset} dataset to train the yolov11-based landmark recognition model. 
%comprising around ${4000}$ images of $34$ real-world landmark instances which were labeled manually. The images were captured using a camera with $2K$-resolution $(2560*1440px)$.
The dataset was split into $70:20:10$ for training, validation, and testing. Image augmentation (brightness adjustment, bounding box rotation \& noise addition, was done on the training portion of the dataset to avoid overfitting problems and to improve the model's robustness. 
% Image color jittering was carried out by adjusting brightness between $-25$ percent and $+25$ percent, bounding box rotation between -15° and +15°,  and bounding box noise addition up to $5$ percent of the pixels. The final dataset comprising $7547$ images including training, validation, and test sets was exported to YOLOv11 Pytorch format.
\subsubsection{MSTlandmarkStereov1 Dataset}
We used a real-world landmark stereo dataset\cite{lanbloc_dataset} for the node-to-landmark distance estimation. We used two identical cameras of $(2560*1440px~@~30~fps)$ resolution %as shown in Fig. \ref{stereo_setup} 
with an adjustable field of view (FoV) to collect this dataset. The cameras were arranged on the metal bar with adjustable baseline(B) from $(10-40)$ centimeters. This dataset was used to perform distance estimation from unknown nodes to the landmark anchors.
%\begin{figure}[!ht]
%\includegraphics[width=0.4\textwidth]{stereo_setup.jpg}
%\centering
%\caption{Stereo Camera Setup for MSTlandmarkStereov1 Collection} 
%\label{stereo_setup}
%\end{figure}

\subsubsection{Trajectory Dataset and  Safe Path}
% \begin{enumerate}
% \item[i)] \textbf{Synthetic Trajectory Dataset:}
\textbf{\\i) Synthetic Trajectory Dataset:}  
Trajectories were generated using initial parameters for acceleration and heading angle over time, calculated for each timestep \(i\) as:  
\[
\begin{split}
% a_{x}(t_i) = -3 \sin\left(\frac{4\pi t_i}{T}\right), \quad  
a_(t_i) = 3 \sin\left(\frac{4\pi t_i}{T}\right), \quad
\theta(t_i) = \frac{\pi}{2} \sin\left(\frac{2\pi t_i}{T}\right)
\end{split}
\]  
With \(N\) timesteps and a timestep size of \(dt = 0.05s\), the total duration was \(T = N \cdot dt\). Each trajectory point was represented by seven state variables: \(p_x, s_x, a_x, p_y, s_y, a_y, \theta\), denoting positions, speeds, accelerations along the x and y axes, and heading direction.

%%%%%%%%%%%%%%%%%%%%%%%%%%%%%%%%%%%%%%%%%%%%%%%%%%%%%%%%%%%%%%%%%%%%%%%%%%%%%%%%%%%%%%%%%%%%%%%
% \item[ii)] \textbf{Real Trajectory Dataset:} 
\textbf{ii) Real Trajectory Dataset:} 
The real trajectory and safe path datasets were obtained to validate the proposed navigation framework. We set a battlefield region of $200*200$ grid space and mapped all the real-world geographical landmarks from MSTlandmarkStereov1 datasets. We randomly assigned 25 obstacles and 25 hazard components within the grid space to simulate the risk and obstacle scenario in the real battlefield as shown in Fig. \ref{battlefield_map}.
Using the landmark dataset, clusters with at least three landmarks within detectable range are selected as the trilateration set for LanBloc localization. Path clusters, defined as sequences of landmark clusters, are used to generate ground truth trajectories from start to end clusters, based on movement through cluster centroids. These trajectories are represented as polylines connecting cluster centroids. The object's states along these trajectories are considered the ground truth trajectory, with trajectory points represented by seven state variables, similar to the synthetic trajectory.

%%%%%%%%%%%%%%%%%%%%%%%%%%%%%%%%%%%%%%%%%%%%%%%%%%%%%%%%%%%%%%%%%%%%%%%%%%%%%%%%%%%%%%%%%%%%%%%
% \item[iii)]\textbf{Safe Path Buffer and Risk Zone:}
\textbf{iii) Safe Path Buffer and Risk Zone:}
% Considering the ground truth trajectory as a central trajectory we obtained the safe path by defining the safety buffer on both sides of the trajectory as shown in the battlefield map in Fig.\ref{battlefield_map}. The safe path is the bounded region where moving objects can navigate safely. The bounded region is considered safe, however, we divided the buffer into four zones of equal width and assigned maximum risk weight to each zone on either side based on the increasing distance from the central trajectory to the buffer boundary to quantify the exposure to the risk.
% Thus, the risk exposure increases as the entity moves towards the buffer boundary and is calculated concerning the deviation from the ground truth trajectory using Eqn \ref{wrs}. we considered $w_1 =2, w_2 =4, w_3 =6, w_4 =8$ as the maximum risk weights assigned to each zone on either side of the central trajectory.
A safe buffer is a bounded region defined on both sides of the ground truth trajectory as in Fig.\ref{battlefield_map}. This bounded region ensures safe navigation and is divided into four equal-width zones. we considered $w_1 =2, w_2 =4, w_3 =6, w_4 =8$ as the maximum risk weights assigned to each zone on either side of the central trajectory. Weighted Risk score anywhere within a zone is calculated based on deviation from the ground truth trajectory using Eqn \ref{wrs} which increases with distance from the central trajectory and quantifies risk exposure.

% \end{enumerate}
% \begin{figure}[!ht]
% \includegraphics[width=0.45\textwidth]{figures/battlefield_scenario_300.png}
% \centering
% \caption{Example showing ground truth trajectory from Cluster 4 to 3} 
% \label{trace_trajectory1}
% \end{figure}

% \begin{figure}[!ht]
% \includegraphics[width=0.45\textwidth]{figures/trajectory_safe_path.png}
% \centering
% \caption{Example Safe Path from Cluster 6 to Cluster 10.} 
% \label{safe_path1}
% \end{figure}
% \end{enumerate}
\vspace{-0.1in}
\subsection{Evaluation Metrics}
% We measured the performance of the proposed approaches with 
Standard evaluation metrics such as Average Displacement Error (ADE), Final Displacement Error (FDE), and Percent Error were used for estimated trajectory evaluation. We also introduced a new metric called \textbf{Average Weighted Risk Score(AWRS)} to evaluate an overall weighted risk exposure of estimated trajectories. To ensure robust and reliable results, the object's motion within each safe path was simulated for 1000 trials, and their average values were calculated.
%Simulating on large sample and averaging their performance metrics provides a thorough, reliable, and statistically significant evaluation of the navigation system’s accuracy and effectiveness in GPS-denied battlefield environments.
% Average Weighted Risk Score (AWRS)
The Average Weighted Risk Score (AWRS) is the mean of the individual Weighted Risk Scores across all trajectory points shown in Eqn. \ref{awrs-metric}.
\begin{equation}
   \text{AWRS} = \frac{1}{N} \sum_{i=1}^{N} \text{WRS}_i 
   \label{awrs-metric}
\end{equation}
Where N  is the total number of trajectory points and $\text{WRS}_i$ is the weighted risk score for the $i^{th}$ point calculated using Eqn. \ref{wrs}.

\begin{equation} 
    ADE = \frac {1}{N}\sum _{t=t_{0}}^{t_{f}} {\|{{\hat {Y}_{(t)}- {Y}_{(t)}}}\|}
    \label{ade}
\end{equation} 

% By simulating many trajectories, the ADE value calculated as the average of these simulations provides a comprehensive measure of the system’s performance, smoothing out random variations and providing a reliable indicator of the typical performance.\\

The ADE is an average of pointwise L2 distances between the estimated and ground truth trajectory calculated using Eqn. \ref{ade}.

The FDE is the L2 distance between the final points of the estimated and ground truth trajectory using Eqn \ref{fde}.
% Averaging FDE over multiple trajectories helps ensure that the final positional accuracy is consistently measured, accounting for different potential paths and variations in movement dynamics.
%\vspace{-0.4in}
% \begin{equation} 
%     FDE = \left |{\left |{{\hat {Y}_{\left ({t_{f}}\right)}- {Y}_{\left ({t_{f}}\right)}}}\right |}\right |
%     \label{fde}
% \end{equation} 

\begin{equation} 
    FDE = {\|{{\hat {Y}_{\left ({t_{f}}\right)}- {Y}_{\left ({t_{f}}\right)}}}\|}
    \label{fde}
\end{equation}
where $\hat{Y}_t$ is the predicted position at time step t and $Y_t$ is the ground truth position.
The \% error of the estimated trajectory indicates a relative error in path length between the predicted trajectory and the actual trajectory calculated using Eqn. \ref{percent_error_eq}.
\begin{equation}
    \text{Percent Error} = \frac{|\text{Est. Length} - \text{True Length} |}{\text{True Length}} \times 100 \%
    \label{percent_error_eq}
\end{equation}
% Calculating the average percent error over 1000 simulations helps to generalize the results and provides a robust measure of the system’s accuracy in estimating the trajectory length.\\
\subsection{Simulation and Experimental Results Discussion}
% \subsection{Result Analysis and Discussion}
\textbf{Baselines:} To evaluate the effectiveness of the proposed navigation algorithm (LanBLoc-BMM), we compared it with various state-of-the-art visual localization and Multi-Source Fusion methods from literature such as DeepLoc-GPS, DeepLoc-EKF-GPS \cite{deepLocalization}, Multiple Sensor Fusion %(IMU/OD/VI/MA) 
\cite{sensor_fusion} using Synthetic and Real-datasets.
The performance results of four algorithms combined with EKF and PF using Synthetic datasets for ground navigation are shown in Table \ref{tab:performance} and using real datasets in Table \ref{tab:performance_realTraj}.
%Comparison of true trajectory, estimated trajectories, and state measured by LanBloc algorithm are shown in Fig.\ref{trace_trajectory1}.
Table \ref{tab:performance} highlights the best overall performance of LanBLoc(EKF), with the lowest ADE, FDE, AWRS, and runtime, while DeepLoc-GPS-EKF (EKF) is a close contender, performing well in all metrics, particularly in FDE and AWRS. DeepLoc-GPS (EKF) also performs well, especially in ADE and FDE; however, it lags in runtime. Sensor Fusion (EKF), while having the highest ADE, shows competitive FDE but is less efficient in computation.
% \vspace{-0.1in}
\begin{table}[ht]
\centering
\caption{Performance across Algorithms %and Filters 
on synthetic trajectory data}

% \begin{tabular}{lccccc}
% \toprule
% \hline
% \textbf{Algorithm} & \textbf{Filter} & \textbf{RMSE} & \textbf{ADE} & \textbf{FDE} &\textbf{T(ms)}\\
% \hline
% \midrule
%         {DeepLocGPS} & KF  & 1.673206 & 0.944307 & 1.127611 &3.2 \\
%                     & \textbf{EKF} & 1.840448 & \textbf{0.606943} & \textbf{0.549535}& 5.02 \\
%                     & PF  & 1.904411 & 1.133528 & 1.761409& 9815 \\
% \hline
% \midrule
%         {DeepLocGPSEKF} & KF  & 1.671124 & 0.925606 & 1.026992&3.0 \\
%                        & \textbf{EKF} & 1.837600 & \textbf{0.562978} & \textbf{0.481784}&4.0\\
%                        & PF  & 1.926583 & 1.486392 & 1.570940 & 11240 \\
% \hline
% \midrule
%         {Sensor Fusion} & KF  & 1.806567 & 1.586412 & 1.432066 &3.0 \\
%                        & \textbf{EKF} & 2.019421 & \textbf{1.354560} & \textbf{0.571424} &\textbf{4.98}\\
%                        & PF  & 2.135636 & 1.915473 & 3.122960 &10760 \\
% \hline
% \midrule
% \textbf{LanBLoc} & KF  & 1.657243 & 0.841321 & 1.054491 &4.02 \\
%                  & \textbf{EKF} & 1.825100 & \textbf{0.497161} & \textbf{0.467504} &\textbf{3.98} \\
%                  & PF  & 1.795693 & 0.947126 & 0.817734 &11610 \\
% \hline
% \bottomrule
% \end{tabular}
% \label{tab:performance}
% \end{table}

\begin{tabular}{lcccccc}
% \toprule
\hline
\textbf{Algorithm} & \textbf{Filter} & \textbf{ADE/FDE (m)} &\textbf{AWRS} &\textbf{Time(ms)}\\
\hline
% \midrule
        {DeepLocGPS} 
        % & KF  & 0.944307 & 1.127611 &3.2 \\
                    & \textbf{EKF}  & 0.6069 / 0.5495&0.3926 &5.02 \\
                    & PF  & 1.1335 / 1.7614&2.1414 &9815  \\
\hline
% \midrule
        {DeepLocGPSEKF} 
        % & KF  & 0.925606 & 1.026992&3.0 \\
                       & \textbf{EKF}& 0.5621/ 0.4818&0.3590 &4.57\\
                       & PF & 1.4864 / 1.5709 &1.7823 &10054 \\
\hline
% \midrule
        {Sensor Fusion} 
        % & KF  & 1.586412 & 1.432066 &3.0 \\
                       & \textbf{EKF} & 1.3546 / 0.5714& 0.3513 &4.90\\
                       & PF & 1.9155 / 3.1229 &2.9387 &10103 \\
\hline
% \midrule
\textbf{LanBLoc-BMM} 
% & KF & 0.841321 & 1.054491 &4.02 \\
                 & \textbf{EKF}  & \textbf{0.4972} / \textbf{0.4675}&\textbf{0.3492}&\textbf{3.98} \\
                 & \textbf{PF} & \textbf{0.9471} / \textbf{0.8177} & \textbf{1.2612} & \textbf{9551} \\
\hline
% \bottomrule
\end{tabular}
\label{tab:performance}
\end{table}

From Table \ref{tab:performance_realTraj}, it is evident that LanBLoc-BMM (EKF) is best regarding accuracy, weighted risk score, and runtime for real trajectory datasets. It consistently offers the most accurate results with reasonable computation times and minimal risk exposure, making it the preferred choice for accurate and efficient localization across different path types. DeepLocGPS(EKF) provides its best accuracy with an ADE/FDE of 0.3041/0.3103, AWRS of 0.2009, and a reasonable runtime of 20.58ms. DeepLocGPSEKF shows the best accuracy with an ADE/FDE of 0.2869/0.2969, AWRS of 0.1906, and a runtime of 18.94 ms. Sensor Fusion(EKF) provides a balance of accuracy (ADE/FDE of 0.2998/0.3115), AWRS of 0.1749, and an acceptable runtime of 22.31ms. The proposed  LanBLoc-BMM (EKF) based navigation again stands out with the best accuracy (ADE/FDE of 0.2393/0.3095), AWRS of 0.1528, and a low runtime of 14.94ms. 
\begin{table}[h!]
\centering
\caption{Average ADE, FDE, WRS, and runtime(ms) values for Real trajectory from three path types [P1, P2, P3] across different algorithms and filter scenarios.}
\begin{tabular}{llccc}
\hline
\textbf{Algorithm} & \textbf{Filter} & \textbf{ADE / FDE(m)}& \textbf{AWRS}&\textbf{Time(ms)} \\ \hline
DeepLocGPS         
% & KF              & 0.3505 / 0.4033                & 15.9   \\  
                    & EKF     & 0.3041 / 0.3103    & 0.2009    & 20.58  \\  
                   &  PF             & 1.6928 / 2.8891          & 4.3586        & 192621  \\ \hline
DeepLocGPSEKF      
% & KF              & 0.3425 / 0.3553                & 14.96  \\  
                   & EKF   & 0.2869 / 0.2969  & 0.1906    &18.94   \\  
                   & PF              & 2.0672 / 7.1665           & 3.6346     & 197250 \\ \hline
Sensor Fusion      
% & KF              & 0.3140 / 0.3505                & 16.75 \\  
                   & EKF     & 0.2998 / 0.3115   & 0.1749    & 22.31 \\  
                   & PF              & 1.6831 / 4.2886          & 5.4116      & 198790 \\ \hline
\textbf{LanBLoc-BMM}            
% & KF              & 0.3122 / 0.3462                & 12.73 \\  
                   & \textbf{EKF}    & \textbf{0.2393 / 0.3095}  & \textbf{0.1528}   & \textbf{14.94}  \\  
                   & PF              & \textbf{1.5162 / 1.4024}  & \textbf{3.1623} & \textbf{185502} \\ \hline
\end{tabular}
\label{tab:performance_realTraj}
\end{table}
% \subsection{}

\begin{figure*}[!ht]
\includegraphics[width=0.94\textwidth]{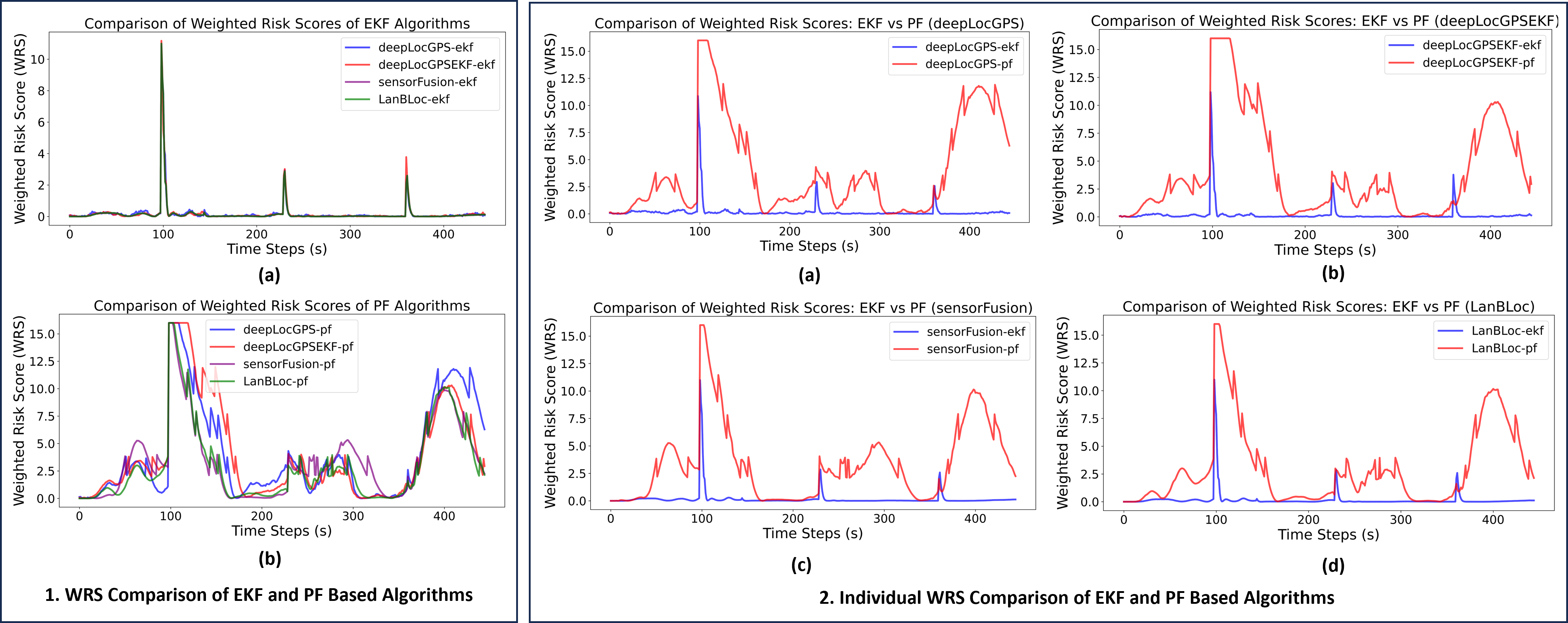}
\centering
\caption{Comparison of Weighted Risk Exposure of EKF-based Algorithms. Fig \ref{wrs-comparisions}.1(a) and \ref{wrs-comparisions}.1(b) show the overall WRS performance of four localization algorithms while using EKF and PF, respectively. The Fig \ref{wrs-comparisions}.2(a),\ref{wrs-comparisions}.2(b),\ref{wrs-comparisions}.2(c) and \ref{wrs-comparisions}.2(d) shows the individual comparison of weighted risk score(WRS) for  EFK and PF integration across each of the four localization algorithms.}
\label{wrs-comparisions}
\end{figure*}

Fig. \ref{wrs-comparisions} shows the overall and individual comparison of WRS of trajectories generated by four different algorithms while integrating EKF and PF separately. Fig. \ref{wrs-comparisions}.1(a) compares the WRS of trajectories generated by four different algorithms when EKF is applied while Fig. \ref{wrs-comparisions}.1(b) shows a similar comparison while PF is applied. Also, Fig. \ref{wrs-comparisions}.1(a),(b),(c), and (d) shows the WRS comparison of each algorithm while applying EKF and PF separately. 

\subsection{Safe Path Navigation using SafeNav-Chull and SafeNav-centroid with Risk-Aware RRT*}
\begin{figure*}[!ht]
\includegraphics[width=0.94\textwidth]{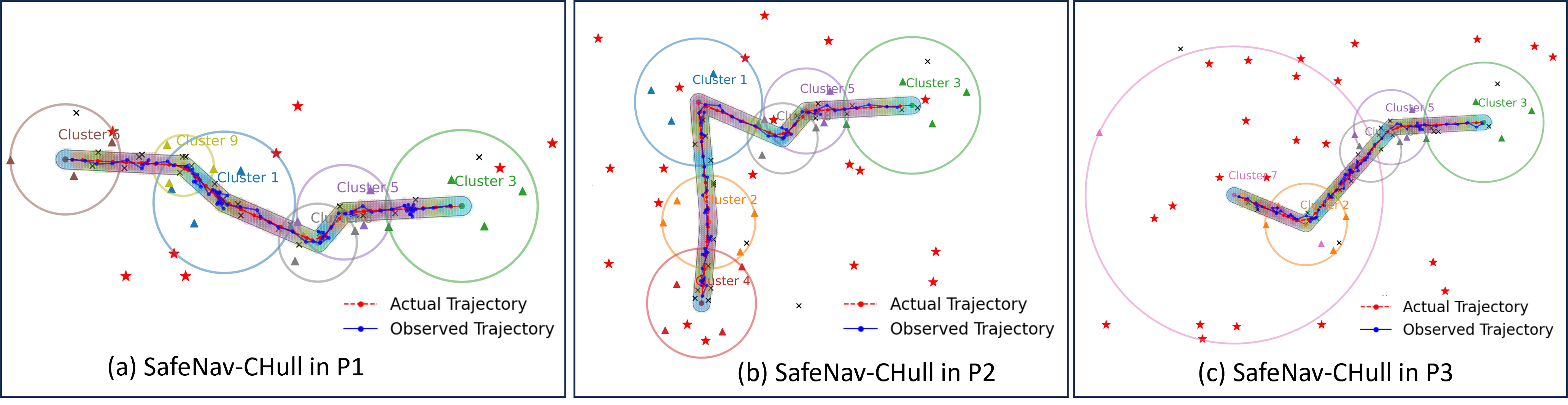}
\centering
\caption{Showing Safe Path Navigation using SafeNav-Chull and RAw-RRT* in three different path types obtained from the combination of different landmark clusters and starting points that guide to the common destination (cluster 3). Fig. \ref{safe_path_chull} (a) shows the path type 1 (P1) from the centroid of cluster 6 to cluster 3. Fig. \ref{safe_path_chull} (b) shows the path type 2  from cluster 4 to cluster 3, and Fig. \ref{safe_path_chull} (c) shows the path type 3 (P3) from cluster 7 to cluster 3. The color coding on each path type represents the convex hull of the path segments. The actual and observed trajectories of moving entities within each path type are shown by red and blue lines, respectively.} 
\label{safe_path_chull}
\end{figure*}

\begin{figure*}[!ht]
\includegraphics[width=0.93\textwidth]{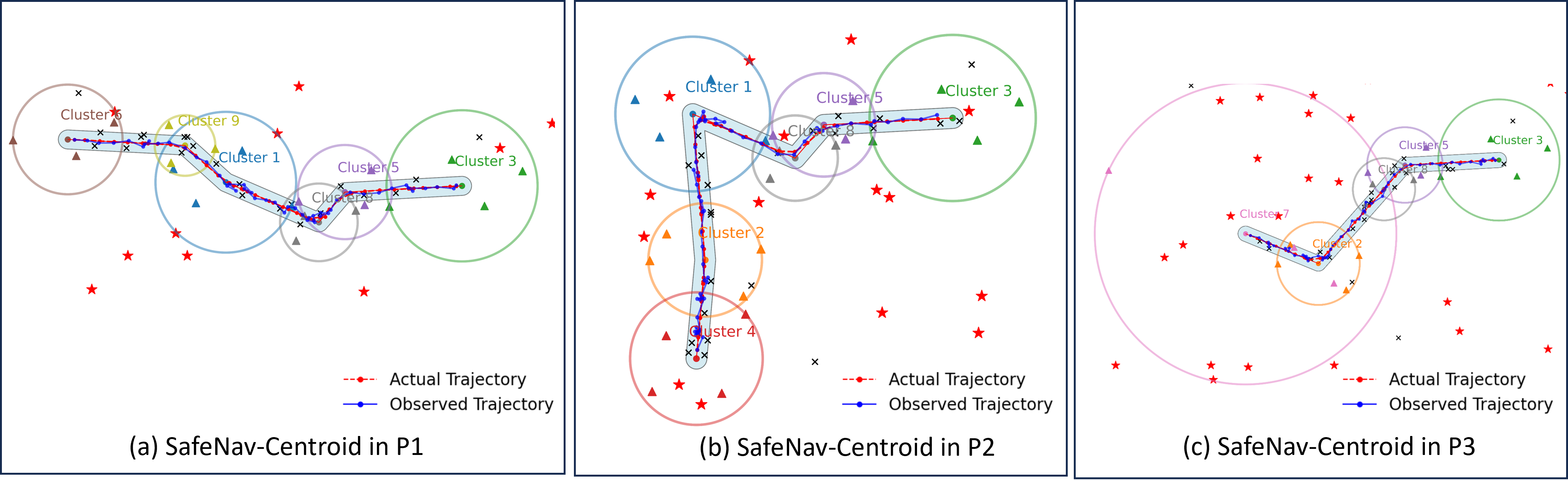}
\centering
\caption{Showing Safe Path Navigation using SafeNav-Centroid and RAw-RRT* in three different path types obtained from the combination of different landmark clusters and starting points that guide to the common destination (cluster 3). Fig. \ref{safe_path_centroid} (a) shows the path type 1(P1) from the centroid of cluster 6 to cluster 3. Fig. \ref{safe_path_centroid} (b) shows the path type 2  from cluster 4 to cluster 3 and Fig. \ref{safe_path_centroid} (c) shows the path type 3 (P3) from cluster 7 to cluster 3. The actual and observed trajectories of moving entities within each path type are shown by red and blue lines, respectively.} 
\label{safe_path_centroid}
\end{figure*}

 % We implemented safety checks and controlled maneuvering approaches in Algorithms \ref{alg_ConvexHull} and \ref{alg_CentroidMethod} for safe navigation applications. 
Next, we design two navigation methods integrating LanBLoc-BMM(EKF) called SafeNav-CHull and SafeNav-Centroid, combining two safety check approaches in Algorithms \ref{alg_ConvexHull} and \ref{alg_CentroidMethod} respectively for safety checks and controlled maneuvering. Then, we integrated the novel RAw-RRT* algorithm with these two navigation approaches for obstacle avoidance and risk minimization within the safe path.
 To test the efficacy of these two navigation algorithms, 
 We ran the navigation trials on three path types drawn from the combination of consecutive cluster members starting from different origins and reaching the same destination. Then, we obtained the trace of the positions that an object traveled along to reach the destination as shown in Fig. \ref{safe_path_chull} and Fig. \ref{safe_path_centroid}.
We simulated 1000 thousand movement trials of an object along each path type and averaged their performance in ADE/FDE, AWRS, and percent error and runtime.
% We calculated the relative error associated with ground truth trajectory and estimated trajectory to obtain the percent error. We calculated the average displacement error (ADE) and Final displacement error (FDE) of estimated trajectories and ground truth trajectories. we also calculated the average weighted risk score (AWRS) of estimated trajectories. 
\subsubsection{Average Percent error in trajectory lengths of two approaches}
Table \ref{percent_error_combined} shows the \% error in the length of the estimated and ground truth trajectory for different path types: $P1$, $P2$, and $P3$. We obtained an average error of $6.51\%$ for SafeNav-CHull.
For the same three path types, we obtained an improved average error of $5.46\%$ for SafeNav-Centroid.

% Overall, the SafeNav-Centroid algorithm, which integrates EKF with
% BBM and LanBLoc outperform SafeNav-CHull in terms of percent error in trajectory estimation. SafeNav-Centroid shows an overall average improvement of 16.12\%  in percent error over the SafeNav-CHull.
%%%%%%%%%%%%%%%%%%%%%%%%%%%%%%%%%%%%%%%%%%%%%%%%%%%%%%%%%%%%%%%%%%%%%%%%%%
% \begin{figure*}[!ht]
% \includegraphics[width=1\textwidth]{figures/safenav-viz.png}
% \centering
% \caption{Showing Safe Navigation using SafeNav-Chull in three different path types of obtained from the combination of different landmark clusters and starting points which guide to the common destination (cluster 3). Fig. \ref{safe_path_chull} (a) shows the path type 1 (P1) from the centroid of cluster 6 to cluster 3. Fig. \ref{safe_path_chull} (b) shows the path type 2  from cluster 4 to cluster 3, and Fig. \ref{safe_path_chull} (c) shows the path type 3 (P3) from cluster 7 to cluster 3. The actual and observed trajectories of moving entities within each path type are shown by red and blue lines, respectively.} 
% \label{safe_path_chull}
% \end{figure*}
%%%%%%%%%%%%%%%%%%%%%%%%%%%%%%%%%%%%%%%%%%%%%%%%%%%%%%%%%%%%%%%%%%%%%%%%%%%%%%%%%%%%
%%%%%%%%%%%%%%%%%%%%%%%%%%%%%%%%%%%%%%%%%%%%%%%%%%%%%%%%%%%%%%%%%%%%%%%%%%%%%%%%%%%%
%%%%%%%%%%%%%%%%%%%%%%%%%%%%%%%%%%%%%%%%%%%%%%%%%%%%%%%%%%%%%%%%%%%%%%%%%%%
%SafeNav-Centroid shows an improvement of $4.03\%$,$4.23\%$, and $4.18\%$ in trajectory estimation for each path type $P1$, $P2$, and $P3$ respectively with an overall average improvement of 4.15\%.\\
From the \% error observed across the three path types, it is evident that the trajectory of the moving object in P2 exhibits higher errors for all algorithms compared to P1 and P3. This increased error is likely due to the drift in the estimated trajectory, which is positively correlated with noise in state variables such as velocity and heading direction. These variables are particularly sensitive to the terrain's characteristics, shape, and length of the path. When an object traverses a path with many turns and challenging terrain, velocity and heading need to be measured more frequently, potentially leading to greater error accumulation than on a path with fewer turns and bends. 
\begin{table}[!ht]
\centering 
\caption{Path length error of estimated trajectories for each path type}
\begin{tabular}{lcccc}
% \toprule
\hline
\multirow{2}{*}{\textbf{Approaches}} & \multicolumn{3}{c}{\textbf{Path Type: Avg. \% Error}} & \multirow{2}{*}{\textbf{Overall Avg.}} \\\cline{2-4}
& P1 & P2 & P3 & \\\hline
% \midrule
\hline
% \hline
% \multicolumn{5}{c}{\textbf{Convex Hull-Based Approaches}} \\
% \midrule
% SafeNav1-CHull & 9.93 & 11.23 & 10.31 & 10.49 \\
\textbf{SafeNav-CHull} & \textbf{5.755} & \textbf{7.42} & \textbf{6.35} & \textbf{6.51} \\
% \midrule
\hline
\hline
% \multicolumn{5}{c}{\textbf{Centroid-Based Approaches}} \\
% \midrule
% SafeNav1-Centroid & 9.35 & 9.82 & 9.65 & 9.61 \\
\textbf{SafeNav-Centroid} & \textbf{5.32} & \textbf{5.60} & \textbf{5.47} & \textbf{5.46} \\
\hline
% \bottomrule
\end{tabular}
\label{percent_error_combined}
\end{table}
%%%%%%%%%%%%%%%%%%%%%%%%%%%%%%%%%%%%%%%%%%%%%%%%%%%%%%%%%%%%%%%%%%%%%%%%%%%%%%%%%%%%%%%%%%%%%%%%%%%%%%%%%%

\subsubsection{Displacement errors and Weighted Risk Scores of estimated trajectories}
Table \ref{displacement_errors_combined} compares  ADE, FDE, and AWRS performances of two navigation algorithms across three different path types (P1, P2, P3). SafeNav-CHull shows higher errors suggesting less accurate trajectory estimation compared to the Centroid-based method. SafeNav-Centroid outperforms SafeNav-Chull, achieving the lowest ADE/FDE rates, indicating the highest accuracy and precision in trajectory estimation. The safeNav-Centroid algorithm is 1.67 times and 3.30 times better, respectively than the ADE and FDE of the Convex-Hull-based approach. The Centroid-based algorithms show better performance in terms of final displacement error than average displacement error which infers that these algorithms allow moving objects to reach precisely close to the intended safe destination.

\begin{table}[!ht]
\centering
\caption{Comparison of ADE/FDE, AWRS, and time for SafeNav-CHull and SafeNav-Centroid}
\begin{tabular}{llllll}
\hline
\textbf{Approaches} & \textbf{Path Type} & \textbf{ADE} / \textbf{FDE} & \textbf{AWRS} & \textbf{Time(ms)} \\
\hline
% \multicolumn{4}{c}{\textbf{Convex Hull-Based Approaches}} \\
\hline
% \multirow{4}{*}{SafeNav1-CHull} 
% & P1 & 5.13 & 6.26 \\ 
% & P2 & 5.11 & 5.86 \\  
% & P3 & 6.45 & 5.86 \\\cline{2-4}
% & \textbf{Average} & \textbf{5.56} & \textbf{5.99} \\
% \hline
\multirow{4}{*}{SafeNav-CHull}
& P1 & 2.73 / 3.16 &1.31 &14.93 \\
& P2 & 2.72 / 3.08 &1.11 &14.84 \\
& P3 & 3.46 / 3.57 &1.71 &15.98 \\\cline{2-5} 
& Average & 2.97 / 3.27 &1.38 &\textbf{15.25} \\
\hline
% \multicolumn{4}{c}{\textbf{Centroid-Based Approaches}} \\
% \hline
% \multirow{4}{*}{SafeNav1-Centroid} 
% & P1 & 1.9  & 1.49 \\
% & P2 & 1.7  & 1.20 \\
% & P3 & 1.85 & 1.26 \\\cline{2-4}
% & \textbf{Average} & \textbf{1.82} & \textbf{1.32} \\
\hline
\multirow{4}{*}{\textbf{SafeNav-Centroid}} 
& \textbf{P1} & \textbf{1.06} / \textbf{0.87} &\textbf{1.01} & 16.74 \\
& \textbf{P2} & \textbf{1.13} / \textbf{1.03} &\textbf{1.01} & 16.94 \\
& \textbf{P3} & \textbf{1.15} / \text{1.06} &\textbf{1.01} & 16.97\\\cline{2-5}
& \textbf{Average} & \textbf{1.11} / \textbf{0.99} &\textbf{1.01}& 16.88 \\
\hline
\end{tabular}
\label{displacement_errors_combined}
\end{table}
\section{Conclusion and Future Works}
In this paper, we first proposed an efficient navigation approach \textbf{LanBLoc-BMM(EKF)} using the LanBLoc as a location measurement model and BMM for motion planning, integrating EKF. Then, we proposed two efficient safe path navigation algorithms, \textbf{SafeNav-Chull} and \textbf{SafeNav-Centroid}, for guiding the mobile troops along the safe and obstacle-free path in GPS-denied battlefield environments. The proposed algorithms combine LanBLoc-BMM(EKF) and geometric properties of the path segments, such as Convex Hull and Centroid, integrated with a novel \textbf{Risk-Aware RRT*} algorithm. The proposed solutions have demonstrated outstanding accuracy in localization and trajectory estimation along defined safe paths. Experimentation and simulation results have validated the effectiveness of the proposed algorithms, showcasing their potential to significantly enhance the safety and operational efficacy of military forces in challenging and hostile environments. This approach addresses the critical vulnerability of GPS reliance, ensuring troops can maneuver safely and efficiently in contested and hazardous areas. 

Our future research will focus on reducing computational demands, improving adaptability to dynamic obstacles and hazard scenarios, and exploring alternative localization strategies for environments lacking identifiable landmarks.
%%%%%%%%%%%%%%%%%%%%%%%%%%%%%%%%%%%%%%%%%%%%%%%%%%%%%%%%%%%%%%%%%%%%%%%%%%%%%%%
%%%%%%%%%%%%%%%%%%%%%%%%%%References%%%%%%%%%%%%%%%%%%%%%%%%%%%%%%%%%%%%%%%%%%%
%-----------------------------------direct bib------------------------------------

% {
%     \small
%     \bibliographystyle{ieeetr}
%     \bibliography{references.bib}
% }

% \bibliographystyle{ieeetr}
% \bibliography{references.bib}
\end{document}